\pdfoutput=1
\documentclass[11pt]{article}

\usepackage[]{acl}
\usepackage{enumitem}
\usepackage{times}
\usepackage{latexsym}

\usepackage[T1]{fontenc}
\usepackage[utf8]{inputenc}

\usepackage{microtype}

\usepackage{amsmath}
\usepackage{enumitem}
\usepackage{float}
\usepackage{graphicx}
\usepackage{multirow}

\usepackage[T1]{fontenc}
\usepackage{dsfont}

\usepackage{mdframed}
\usepackage{booktabs}
\usepackage{graphicx}
\usepackage{subcaption}

\usepackage{enumitem}
\usepackage{xcolor}
\usepackage{soul}
\colorlet{soulred}{red!30}
\sethlcolor{soulred}%

\usepackage{amsmath,amsthm,amsfonts,amssymb,bm}
\usepackage{framed}
\usepackage{varioref}
\usepackage{float}
\usepackage{multirow}
\usepackage{dashrule}
\usepackage{url}
\usepackage{pifont}% http://ctan.org/pkg/pifont
\usepackage{helvet}

\newcommand{\bfsec}[1]{{\noindent\textbf{#1}}}

\usepackage[most]{tcolorbox}
\usepackage{lipsum}
\usepackage{algorithm}

\definecolor{green}{rgb}{0.1,0.1,0.1}
\definecolor{chocolate}{HTML}{D2691E}
\definecolor{maroon}{HTML}{A00000}
\definecolor{indigo}{HTML}{4B0082}
\definecolor{green}{HTML}{008000}
\definecolor{cadmiumgreen}{rgb}{0.0, 0.42, 0.24}

\usepackage{amssymb}% http://ctan.org/pkg/amssymb
\usepackage{pifont}% http://ctan.org/pkg/pifont

\makeatletter
\newcommand*\myfontsize{%
  \@setfontsize\myfontsize{8}{9}%
}
\makeatother

\makeatletter
\newcommand*\mysmallfontsize{%
  \@setfontsize\mysmallfontsize{7.4}{8.4}%
}
\makeatother

\usepackage{adjustbox}

\newcommand{\myskip}[1]{}

\usepackage{makecell}
\usepackage{algpseudocode}
\tcbset{
    appendixbox/.style={
        colframe=black,       % Border color
        colback=gray!20,      % Background color
        boxrule=1mm,          % Border thickness
        arc=4mm,              % Rounded corners
        left=2mm,             % Left padding
        right=2mm,            % Right padding
        top=2mm,              % Top padding
        bottom=2mm,           % Bottom padding
        breakable,            % Allow box to break across pages
    }
}

\title{\textsc{LDC}: \underline{L}earning to Generate Research Ideas with \underline{D}ynamic \underline{C}ontrol
}

\author{%
    Ruochen Li$^1$, Liqiang Jing$^1$, Chi Han$^2$, Jiawei Zhou$^3$,  Xinya Du$^1$ \\
    $^1$University of Texas at Dallas \quad $^2$UIUC  \quad $^3$ Stony Brook University \\
    \texttt{\{ruochen.li, liqiang.jing, xinya.du\}@utdallas.edu} \\ 
    \texttt{chihan3@illinois.edu}\\ \texttt{jiawei.zhou.1@stonybrook.edu} \\
}
\begin{document}
\maketitle

\begin{abstract}
\vspace{-1mm}
Recent advancements in large language models (LLMs)
have demonstrated their potential in automating the scientific research ideation.
Existing approaches primarily focus on prompting techniques, often producing ideas misaligned with expert standards -- novelty, feasibility, and effectiveness, which are widely recognized by the research community as the three key subdimensions of high-quality ideas.
Also, balancing these dimensions remains challenging due to their inherent trade-offs.
To address these limitations, we propose the first framework that employs a two-stage approach combining Supervised Fine-Tuning (SFT) and controllable Reinforcement Learning (RL) for the task. 
In the SFT stage, the model learns foundational patterns from pairs of research papers and their corresponding follow-up ideas.
In the RL stage, multi-dimensional reward models guided by fine-grained feedback evaluate and optimize the model across key dimensions.
During inference, dimensional controllers coordinated by a sentence-level decoder enable dynamic context-aware steering of the idea generation process.
Our framework provides a balanced approach to research idea generation, achieving high-quality outcomes in the experiment by dynamically navigating the trade-offs among novelty, feasibility, and effectiveness.
\end{abstract}

\section{Introduction}
\begin{figure}[t]
    \centering
    \includegraphics[width=\linewidth]{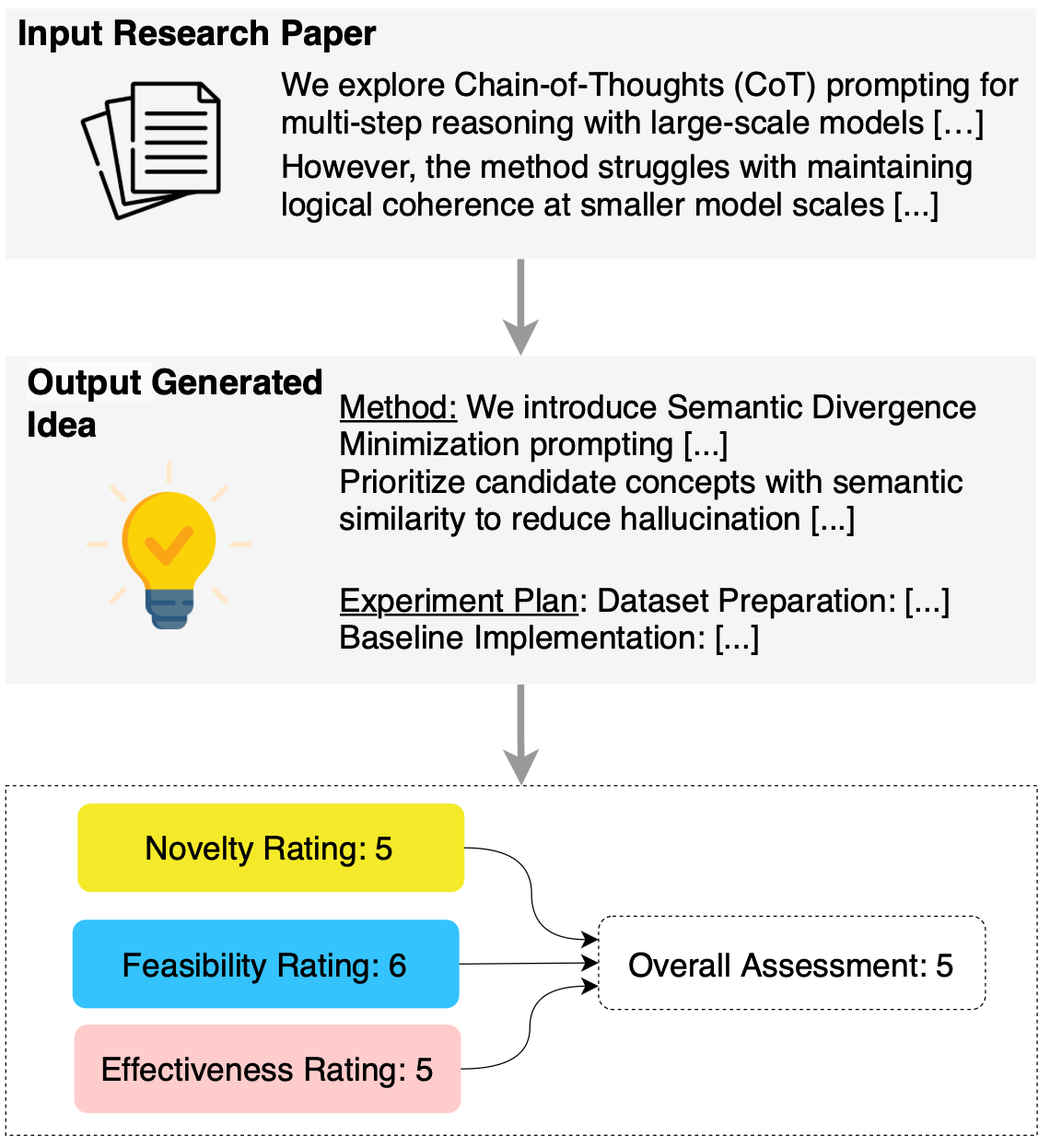}
    \caption{Research idea generation from research papers. Each idea is measured across the dimensions of novelty, feasibility, and effectiveness.}
    \vspace{-4mm}
    \label{fig:fig1}
\end{figure}
\vspace{-2mm}

Typically, a well-developed scientific research idea (or hypothesis\footnote{In this paper, research idea and hypothesis are used interchangeably.}) consists of a \textit{methodology} and an \textit{experiment plan}, as illustrated in Figure~\ref{fig:fig1}.
The \textit{methodology} introduces the novel concept or approach, while the \textit{experiment plan} provides a structured guide for its validation.
Formulating such research ideas is fundamental to the research process.
Traditional methods, which rely heavily on human intuition and experience, are often time-consuming and prone to biases.
In contrast, automated research idea generation systems can swiftly synthesize vast data and insights, uncovering novel connections beyond human researchers.
Recent work using LLM-based agents has demonstrated their potential for generating and validating innovative ideas~\cite{baek2024researchagent, bornstein2024hypothesiscraft}.
Despite the notable progress, these efforts primarily rely on pre-trained models without task-specific learning, which restricts the full exploitation of optimizing the generated content toward scientific expert standards.

Recent studies and expert interviews show that novelty, feasibility and effectiveness are widely recognized by the research community as the three key subdimensions of high-quality research ideas~\cite{si2024llmsgeneratenovelresearch,baek2024researchagent}.
Specifically, novelty reflects the originality of the idea; feasibility assesses its practicality given current resources and constraints; and effectiveness measures the likelihood that the idea will achieve its intended outcomes.
These fine-grained metrics, alongside the overall rating, can help evaluate ideas and guide generation through optimization techniques such as reinforcement learning (RL); more specifically, Reinforcement Learning from Human Feedback (RLHF) can be used to optimize LLM toward scientist standards \cite{ouyang2022training}.
Despite these advancements, existing approaches cannot tackle the complex interdependence and inherent restrictions among these dimensions.
One notable challenge identified is to reveal the inevitable \emph{innovation-feasibility trade-off}~\cite{Yang2023LargeLM, si2024llmsgeneratenovelresearch}: highly novel ideas often lack feasibility, while overly feasible ideas tend to limit the scope for groundbreaking discoveries.
Optimizing idea generation towards each of the key dimensions while achieving a balanced trade-off remains a critical yet unresolved question.

To address this, we propose a framework to improve the intrinsic capabilities of LLMs on generating research ideas. It dynamically adjusts the emphasis on key dimensions of the research idea to achieve high overall quality through a two-stage training process: SFT and controllable RL.
In the SFT stage, the idea proposer learns foundational patterns by training on pairs of research papers and corresponding follow-up ideas.
In the RL stage, we employ multidimension reward modeling as a real-world assessment approximation~\cite{DBLP:conf/nips/WuHSDSASOH23}. Reward models, trained on automatically obtained fine-grained feedback from review data, score each dimension--providing detailed guidance for model refinement.
To enable precise and adaptive control, we introduce dimensional controllers, trained alongside the RL process, which adjust the generation to prioritize specific dimensions when necessary.
This is done at inference time by a sentence-level decoder that dynamically adjusts the weights of controllers, ensuring context-aware emphasis--such as prioritizing novelty in the method part and feasibility in the experiment planning.
Together, these mechanisms, guided by feedback signals from the reward models, result in more balanced and high-quality idea generation.

Our contributions are summarized as follows: 
\begin{itemize}[noitemsep, topsep=0pt]
\item We introduce a two-stage fine-tuning framework for LLM-based research ideation, which dynamically optimizes idea generation towards novelty, feasibility, and effectiveness.
\item We introduce a dynamic decoding to address interdependent subdimensions such as novelty and feasibility.
\item We leverage automatically collected real-world data to train reward models that provide automated, fine-grained feedback aligned with expert evaluations.
\item We conduct comprehensive evaluations, which demonstrate the effectiveness of our method for optimized and controllable research idea generation.
\end{itemize}
\vspace{-1mm}

\section{Related Work}
\begin{figure*}[h]
\vspace{-2mm}
    \centering
    \includegraphics[width=\linewidth]{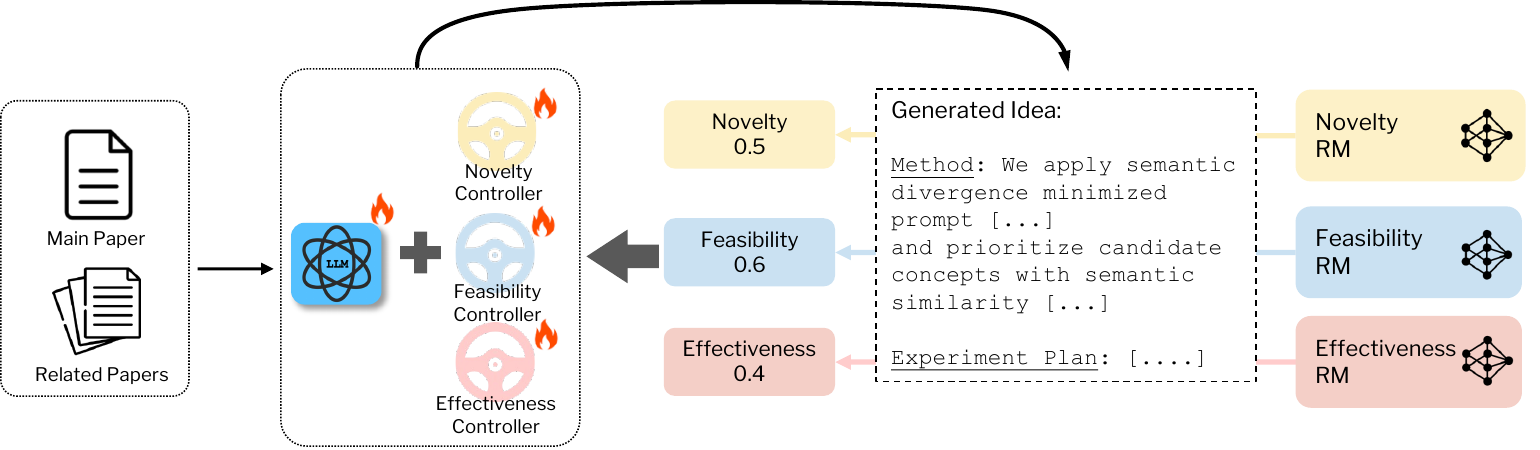}
    \vspace{-5mm}
    \caption{The learning framework with dynamic control across 3 dimensions. Generated research ideas are assessed by corresponding reward models, which provide scores for each dimension. These scores guide the fine-tuning process during reinforcement learning, optimizing both the idea proposer and the corresponding dimensional control parameters to enhance the quality of idea generation. Fires denote weight changes during the process.
%     % dynamic adjusting for the intended balance between them.
    }
    \label{fig:fig2}
    \vspace{-3mm}
\end{figure*}
    \vspace{-1mm}

% \begin{figure*}[h]
%     \centering
%     \includegraphics[width=\linewidth]{figures/fig2.pdf}
%     \caption{The learning framework with dynamic control across 3 dimensions. Generated research ideas are assessed by corresponding reward models, which provide scores for each dimension. These scores guide the fine-tuning process during reinforcement learning, optimizing both the idea proposer and the corresponding dimensional control parameters to enhance the quality of idea generation.
%     % dynamic adjusting for the intended balance between them.
%     }
%     \label{fig:fig2}
%     \vspace{-2mm}
% \end{figure*}

\vspace{-0.5mm}
\textbf{NLP for scientific discovery.}
% Several prior works explored methods to improve idea generation, such as iterative novelty boosting~\cite{Wang2023SciMONSI}, multi-agent collaboration~\cite{Baek2024ResearchAgentIR}, and multi-module retrieval and revision~\cite{Yang2023LargeLM}. 
% While some of them share similar components as our ideation agent, these works focus on improving the idea generation methods over vanilla prompting baselines, without comparisons to any human expert baselines. 
% Beyond ideation, another line of work uses LLMs for executing experiments by generating code given the research problems~\cite{Huang2023MLAgentBenchEL,Tian2024SciCodeAR}, or combining idea generation with code generation to directly implement AI-generated ideas~\cite{AIScientist,Li2024MLRCopilotAM}. 
% These works either use automatic evaluation on a pre-defined set of problems and benchmarks, setting a constrained problem space; or rely on proxy metrics like LLM evaluators, which are often unreliable. 
NLP techniques have significantly advanced scientific discovery by enabling researchers to manage extensive literature, identify knowledge gaps, and analyze trends effectively~\cite{raghu2020survey, hope2021scisight}.  Models such as SciBERT~\cite{beltagy2019scibert} and BioBERT~\cite{lee2020biobert} pre-trained on scientific materials have enhanced these abilities by improving performance on fundamental tasks.
Recent developments in LLMs have extended their utility to creative and generative tasks in scientific research. For example, LLMs have been employed to formulate research questions, generate hypotheses, draft research proposals, and even outline experimental designs~\cite{brown2020language, zhong2023goal, qi2023large, Yang2023LargeLM, wang2024scimonscientificinspirationmachines}.
Several prior works have specifically explored methods to enhance idea generation. Approaches such as iterative novelty boosting~\cite{Wang2023SciMONSI}, multi-agent collaboration~\cite{baek2024researchagent}, and multi-module retrieval and revision~\cite{Yang2023LargeLM} have been proposed to advance ideation capabilities beyond baseline prompting methods.
Beyond ideation, other researchers leverage LLMs for automating experimental workflows. Works like MLAgent~\cite{huang2024mlagentbench} and SciCode~\cite{Tian2024SciCodeAR} use LLMs to generate code for executing research experiments, while AI Scientist~\cite{AIScientist} and MLR-Copilot~\cite{Li2024MLRCopilotAM} combine idea generation with code implementation to directly test AI-generated concepts. However, these approaches are often limited to constrained problem spaces or rely on proxy metrics for evaluation, such as LLM-based scoring, which can be inconsistent and unreliable.
\vspace{-2mm}

\paragraph{Fine-tuning LLM with RL.}
RLHF has shown success in diverse NLP tasks~\cite{christiano2017deep, stiennon2020learning, ouyang2022training}, including text summarization~\cite{ziegler2019fine}, instruction following~\cite{ouyang2022training}, and question answering~\cite{nakano2021webgpt}. While most works focus on optimizing a single holistic reward combining multiple objectives, recent efforts have explored rewards modeling for multiple specific attributes, such as reasoning or ethical considerations~\cite{glaese2022improving, uesato2022helpful}. In this work we investigate fine-grained rewards for the more challenging problem of optimizing multiple dimensions.

\vspace{-1.5mm}

\section{Method}
We introduce a scientific idea proposer with multi-dimension feedback, which consists of two stages: supervised fine-tuning stage, and reinforcement learning stage that has three components: reward modeling, multi-dimension reward augmented controllable reinforcement learning, and decoding.

\subsection{Overview}
Suppose we have a training set  $\mathcal{D} = \{X_i, Y_i\}_{i=1}^N$, where $X_i$ and $Y_i$ are research paper and idea, respectively. 
Then we fine-tune the language model $\mathcal{M}$ with the training set. 
Thereafter, we collect a reward training set  $\mathcal{D}_r = \{(X_i^r, Y^n_i, Y^f_i, Y^e_i)_{i=1}^N\}$, where $X_i$ include the textual content of research paper and research idea, and $Y^n_i, Y^f_i, Y^e_i$ are the labels which show the scores of novelty, feasibility, and effectiveness of research idea.
We could utilize this training set to train three reward models as follows,
\vspace{-2mm}

\begin{equation}
 \left\{
 \begin{aligned}
&F_n = \mathcal{R}_{n} (X_i^r, Y^n_i|{\Theta}_{n}),\\
&F_f= \mathcal{R}_{f} (X_i^r, Y^f_i |{\Theta}_{f}), \\
&F_e= \mathcal{R}_{e} (X_i^r, Y^e_i  |{\Theta}_{e}).
\end{aligned}
 \right.
\end{equation}

where $\Theta_{n/f/e}$ is the parameters of the reward model $\mathcal{R}_{n/f/e}$. 
$\mathcal{R}_{n/f/e}$ denotes the reward models that aim to score the novelty, feasibility, and effectiveness of the research idea. $F_{n/f/e}$ are values from reward models.  
Then, we use a set of $N_f$ research papers $ \{{P_i}\}_{i=1}^{N_f}$ as input to the language model to generate research ideas, which are assessed with reward models based on three criteria.
Finally, we conduct reinforcement learning on the language model as,
\begin{equation}
{H}=\mathcal{M}(P |{\Theta}_{m}, \Theta_{n}, \Theta_{f}, \Theta_{e}),
\end{equation}
where $\Theta_{m}$ denotes the final optimized parameters of the language model $\mathcal{M}$.
% During which the dimensional controllers are jointly trained to improve its ability to generate high-quality research ideas with fine-grained control at inference time.
During this process, three dimensional controllers are trained jointly with the language model to enable fine-grained control at inference time. 

\subsection{Supervised Fine-Tuning}
To improve model training stability in RL~\cite{DBLP:journals/corr/abs-2406-10305}, we also introduce the supervised fine-tuning stage.
The goal of this stage is to introduce the model with the general task format and stabilize the subsequent RL stage.
Therefore, the training data at this stage does not need to achieve high scores in terms of the metrics, which will later be optimized through the fine-grained RL.

% \textbf{Data Collection.}
\bfsec{Data Collection.}
\label{sec:data_collect}
To conduct a supervised fine-tuning stage, we need to collect a set of research papers $\{X_i\}_{i=1}^N$, which we name as supporting papers, and a collection of research ideas $\{Y_i\}_{i=1}^N$, each inspired by a corresponding supporting paper. 
To collect high-quality research ideas, we first collect papers from ICLR 2023 and 2024.
As a top-tier conference in the field of machine learning that covers diverse domains and topics, ICLR is renowned for its cutting-edge research and high-quality technical discussions,  making it an ideal source for this purpose.
We sample 1,000 instances of papers $\{p\}$, and then utilize the LLaMA with a prompt (detailed in Appendix~\ref{appendix:extract_idea}) to extract the research idea $y$ from the sampled paper $p$ as the golden output.
% Paper ICLR 2023 2024 1000 samples
To extract the one corresponding supporting paper $X_i$, i.e. the input of each extracted research idea $Y_i$, for each output, we select the one most significant supporting paper from all related works $\hat{x_1}, \hat{x_2}...,\hat{x_n}$ by prompting LLaMA with the abstract and introduction section of $p$, together with the citation counts of $\hat{x_1}, \hat{x_2}...,\hat{x_n}$ within the sampled paper $p$.
For all extraction, we use LLaMA3 70B to ensure high-quality results.

% \textbf{Fine-Tuning.}
\bfsec{Fine-Tuning.}
Based on the collected training set $\mathcal{D} = \{X_i, Y_i\}_{i=1}^N$, we fine-tune the language model $\mathcal{M}$ as follows,
\vspace{-2mm}
\begin{equation}
    \mathcal{L}_{sup} = CE(Y, \hat{Y})
\end{equation}
\vspace{-5mm}

\noindent where $CE(\cdot)$ denotes the cross-entropy loss and $\hat{Y}$ is the predicted research idea from $\mathcal{M}$, formulated as $\hat{Y}=\mathcal{M}(X)$. 
\vspace{-2mm}

\subsection{Reward Modeling}

Researchers mainly consider three aspects when they devise research ideas: novelty, feasibility, and effectiveness.
These aspects are also used in the review process as fine-grained dimensions of research ideas besides an  overall quality.
Therefore, we train three distinct reward models to score the generated idea in reinforcement learning, each corresponding to one of the quality dimensions. 

% \textbf{Data Collection}

% ICLR 2023 2024 TITLE ID

% Semantic ScholaR RELATED WORK (Top-3 related work (title + abs) title by pass checkpoint as input)

% Arxiv -> download paper
% pdf source -> clean section abs method experiment
% 23 novelty score 24 none
% 24 -> prompt review + paper abs -> prompt novelty review -> GPT
% feasibility -> prompt research hypotheses (abs + method generate prompt from the original paper + criteria + example) title experiment

% effectiveness

% \textbf{Multi-dimension Feedback Collection.}
\bfsec{Multi-dimension Feedback Collection.}
To train reward models, we need to collect three kinds of feedback. 
% Therefore, to optimize the idea-generation agent, we need to collect three kinds of feedback. 
% By collecting this feedback, we could further train three reward models for scoring the hypotheses generated by an agent. 
Similar to the supervised fine-tuning stage, we use the papers from ICLR\footnote{\url{https://iclr.cc/}} and NeurIPS\footnote{\url{https://neurips.cc/}} due to their availability and high quality.
Specifically, we collect the review data from OpenReview, and we extract the research ideas also with prompting.
For the Novelty score of the research ideas in the year 2023, we could use the novelty score from the review directly.
As for those in the year 2024, we prompt Llama3 to get novelty scores since they don't provide direct ratings (see Appendix~\ref{appendix:prompt-nov} for prompts).
% and paper content is scraped from the Semantic Scholar\footnote{\url{xxx}.} and arXiv APIs\footnote{\url{xxx}.} and then cleaned up with regular expression to extract corresponding sections. 
Similarly, since there is no feasibility score or effectiveness score in the review, we prompt Llama3 to get scores for every research idea.
Feasibility score is mainly based on the experiment setup and method sections, taking into account factors such as dataset size, model complexity, and relevant review comments, while Effectiveness score is derived primarily from the experimental results and corresponding review comments.
For all extraction with Llama3 we use the 70B API.
The detailed Scoring Criteria for Novelty, Feasibility, and Effectiveness are outlined in Appendix~\ref{appendix:def}.

Notably, all the collected novelty, feasibility, and effectiveness are subsequently normalized to a 0-1 scale for training.  
% We show all prompts in the scoring process in Appendix \ref{}.

% \textbf{Reward Model Training.}
\bfsec{Reward Model Training.}
We select an LLM as the backbone of reward models. To make the model predict the score for each dimension, we add a Multi-Layer Perceptron as follows,
\begin{equation}
\small
 \left\{
 \begin{aligned}
& \mathbf{F}_{n/f/e} = \mathcal{A}_{_{n/f/e}} (X^r),\\
&\hat{F}_{n/f/e}= \mathcal{C}_{n/f/e} (\mathbf{F}_{n/f/e}), 
\end{aligned}
 \right.
 \end{equation}
where $\mathcal{C}_{n/f/e}$ are MLPs which can output score for each dimension.  $\mathcal{A}_{n/f/e}$ is the LLM backbone.
Each reward model takes the generated idea as input and outputs a score $F_{n/f/e}$  between 0 and 1, representing its evaluation of novelty, feasibility, or effectiveness. 
To optimize the reward models, we utilize cross-entropy loss as follows,
\begin{equation}
\small
    \mathcal{L}_{n/f/e} = CE(\hat{F}_{n/f/e}, F_{n/f/e}),
\end{equation}
where $ F_{n/f/e}$ is the ground-truth label.

\subsection{Multi-dimension Reward Augmented Controllable Reinforcement Learning}
\label{sec:rl}
In this stage, we fine-tune the research idea proposer with controllable steering through reinforcement learning (Figure~\ref{fig:fig2}), refining the model based on feedback across three dimensions: novelty, feasibility, and effectiveness.
\begin{figure*}[t]

    \centering
    \includegraphics[width=0.9\linewidth]{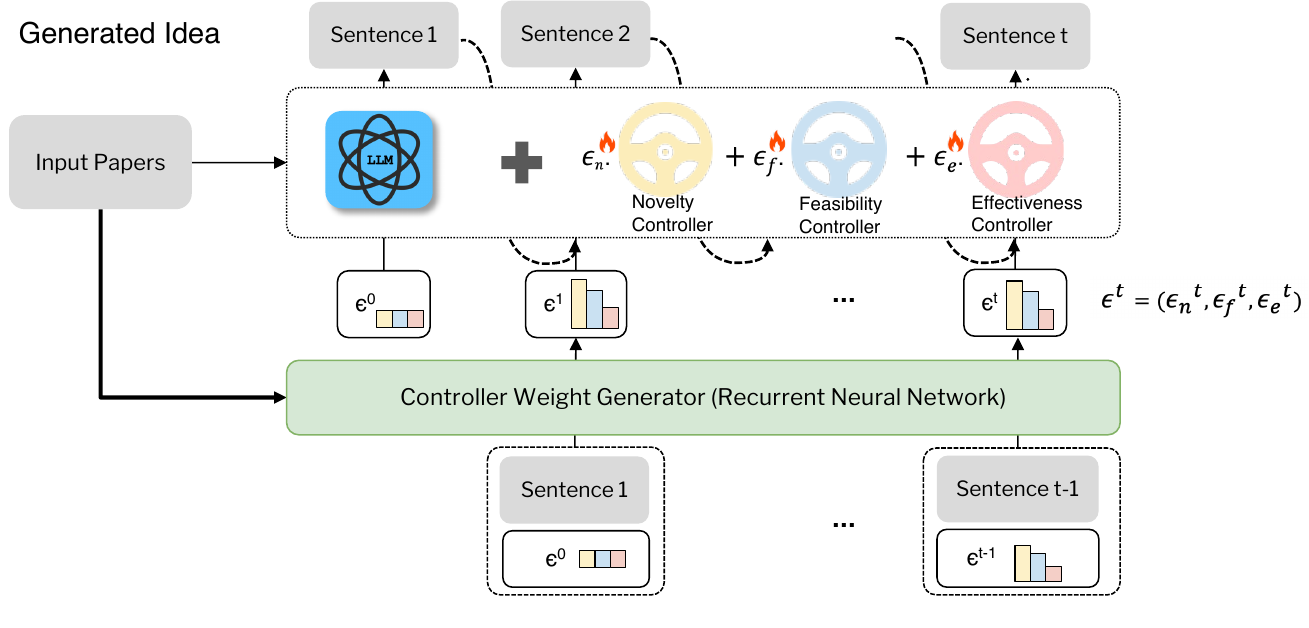}
    \vspace{-4mm}
    \caption{Decoding RNN dynamically steers the dimensions for a balanced and context-aware generation. The process starts with $\epsilon^{0}$ and predicts the control weights for the next sentence condition on the generated context.}
    \label{fig:decode}
    \vspace{-4mm}
\end{figure*}\\
\textbf{Dimensional Controllers.}
Inspired by the existing work \cite{han2024word}, we introduce the dimensional controllers of the novelty, feasibility, and effectiveness of the generated idea, as these dimensions often exhibit interdependency and trade-offs.
We achieve this by adding additional control parameters (i.e. the steers) as follows,
\begin{equation}
\small
 \left\{
 \begin{aligned}
&  \mathbf{M}^l_n = \mathbf{M}_l + \epsilon_n \mathbf{W}_n \mathbf{M}_l,\\
&\mathbf{M}^l_f = \mathbf{M}_l + \epsilon_f \mathbf{W}_f \mathbf{M}_l,\\
&\mathbf{M}^l_e = \mathbf{M}_l + \epsilon_e \mathbf{W}_e \mathbf{M}_l,
\end{aligned}
 \right.
 \end{equation}
where $\mathbf{M}_l$ represents the output of $l$-th layer in the LLM. $\epsilon_n$, $\epsilon_f$, and $\epsilon_e$ are the hyper-parameters for controlling novelty, feasibility, and effectiveness. $\mathbf{W}_n$, $\mathbf{W}_f$, and $\mathbf{W}_e$ are learnable parameters. 
In the training stage, we set all $\epsilon_n$, $\epsilon_f$, and $\epsilon_e$ as 1. By this, we use $\mathbf{M}^l_{n/f/e}$ to replace the original output of the $l$-th layer. We denote the parameters for each resulting model as $\Theta_n = \{\Theta_{LLM}, \Theta_{\epsilon_n \mathbf{W}_n \mathbf{M}_l}\}$, $\Theta_f = \{\Theta_{LLM}, \Theta_{\epsilon_f \mathbf{W}_f \mathbf{M}_l}\}$ and $\Theta_e = \{\Theta_{LLM}, \Theta_{\epsilon_e \mathbf{W}_e \mathbf{M}_l}\}$.
Each controller thus acts as a layer-wise multiplicative perturbation of the hidden states: with $\epsilon=0$ the model reduces exactly to the underlying RL-tuned policy, and $\epsilon$ scales the projection of $\mathbf{M}_l$ onto the learned dimension-specific subspace, so the three controllers can be composed at inference time without retraining.
We further discuss in Appendix~\ref{appendix:steer_theory} why the three controllers are trained independently rather than against a single scalarized reward.

% \textbf{Reward.}
\bfsec{Reward.}
Specifically, we get all three kinds of rewards for each research idea based on the well-trained reward model. We define $r_n$, $r_f$, and $r_e$ as the novelty, feasibility, and effectiveness rewards for the research idea.  
Then we have a reward function for each dimension of the research idea at timestep $t$ as follows, 
\begin{equation}
\small
 \left\{
 \begin{aligned}
&  r^n_t = - {\sum_{i=1}^{t}{\mathbb{I}(i=K)w_l r_n} },\\
&r^f_t = - {\sum_{i=1}^{t}{\mathbb{I}(i=K)w_l r_f} },\\
&r^e_t = - {\sum_{i=1}^{t}{\mathbb{I}(i=K)w_l r_e} },
\end{aligned}
 \right.
 \end{equation}
where $K$ is the token length of the research idea. $t$ is the timestep. $\mathbb{I}(\cdot)$ is the indicator function. $w_l$ is a weight assigned to rewards. Thereafter, we utilize the {PPO algorithm} \cite{DBLP:journals/corr/SchulmanWDRK17} to train the model following the existing  work~\citep{jing2024fgaifaligninglargevisionlanguage}. 
More details are in Appendix~\ref{appendix:ppo}.

% \subsection{Decoding}

\subsection{Decoding}
In this part, we devise two decoding methods for the inference stage. 

% \textbf{Naive Static Decoding.}
\bfsec{Naive Static Decoding.}
In this decoding method, we set $\epsilon_n$, $\epsilon_f$, and $\epsilon_e$ as fixed values for the steers. 
To achieve a high score over novelty, feasibility, and effectiveness, we set all $\epsilon_n$, $\epsilon_f$, and $\epsilon_e$ as $1$, because we set them as $1$ in the training stage for maximum novelty, feasibility, and effectiveness.

% \textbf{Goal-driven Dynamic Decoding.}
\bfsec{Goal-driven Dynamic Decoding.}
The goal of achieving a good research idea is not only to improve the result of a certain dimension but also to consider the overall quality. For example, very high degree of novelty may result in low effectiveness~\cite{si2024llmsgeneratenovelresearch, yang2024large}, while different parts of a research idea, such as method and experiment planning, may require varying levels of focus on novelty and feasibility.
Therefore, how to balance novelty, feasibility, and effectiveness during inference is important.
To achieve this, we utilize a recurrent neural network (RNN)~\cite{sherstinsky2020fundamentals} to predict the steer value $\epsilon_n$, $\epsilon_f$, and $\epsilon_e$~(Figure \ref{fig:decode}), as RNN is good at sequence-level prediction. 
% while it introduces minimal inference time overhead in our task.

To optimize the RNN for steer values prediction, we first collect 1,000 high-quality research ideas generated with Idea Proposer  (scoring above 8 overall). 
Thereafter, we get the corresponding controller weights using our three reward models for each sentence of the high-quality research idea.
Specifically, we feed each sentence in the research idea into our reward models to get the rewards as $\hat{r}_n$, $\hat{r}_f$, $\hat{r}_e$.
Furthermore, we normalize the 
reward to reflect the controller-weight ratios between three controllers, as well as the absolute scale of each controller weight from 0.0--5.0. The corresponding steer values of each sentence $s_t$ are computed as:
% $\hat{\epsilon}_{n/f/e} = (\hat{r}_{n/f/e} - r_{\min}})/ ({r_{\max}} - r_{\min}) \cdot \epsilon_{\max}$.
$ \hat{\epsilon}_{n/f/e} = (\hat{r}_{n/f/e} - r_{\min}) / (r_{\max} - r_{\min}) \cdot \epsilon_{\max}$
% where $r_{{n/f/e}_{min}}$ and $r_{{n/f/e}_{max}}$ are the minimum and maximum value for all rewards and $w'$ is the maximal controller weight, which is 5 in our case.
where $r_{\min}$ and $r_{\max}$ denote the minimum and maximum value for all rewards, and $\epsilon_{\max}$ is the maximal controller weight.
After the data collection, we can use the pair $(S^t, \hat{\epsilon}_{n/f/e}^{t})$ to train the model:
\begin{equation}
% \small
    \mathcal{L}_{rnn} = CE(RNN(S^{<t}),\hat{\epsilon}_{n/f/e}^{t}),
\end{equation}
where $S^{<t}$ is the preceding $t{-}1$ sentences generated in the research idea. Afterward, we use the trained RNN to predict the controller weights of the next sentence $\epsilon^{t} = (\epsilon^{t}_n, \epsilon^{t}_e, \epsilon^{t}_f)$ based on $\epsilon^{t-1}$ and previous sentence.

Finally, during inference, we apply the controller weights by adding them on top of the LLM last layer embedding $\mathbf{M}_l$ to steer the generation:
\begin{equation}
% \small
    \mathbf{M}^l_n = \mathbf{M}_l + \epsilon_n \mathbf{W}_n \mathbf{M}_l + \epsilon_f \mathbf{W}_f \mathbf{M}_l + \epsilon_e \mathbf{W}_e \mathbf{M}_l
\end{equation}

\vspace{-1.5mm}

\vspace{-2mm}
\section{Experiment}
\vspace{-1mm}

\subsection{Dataset}
\vspace{-0.5mm}

% \begin{figure}[h]
%     \centering
%     \includegraphics[width=0.7\linewidth]{figures/stats_barplot.png}
%     \caption{Rating distribution statistics of our dataset.}
%     \label{fig:fig_stat}
% \end{figure}

% \begin{figure}[h]
%     \centering
%     \includegraphics[width=0.7\linewidth]{figures/topic-new.png}
%     \caption{Top 10 topic distribution of our dataset.}
%     \label{fig:fig_stat2}
% \end{figure}
% \begin{figure}[h]
%     \centering
%     \begin{subfigure}[t]{0.48\linewidth}
%         \centering
%         \includegraphics[width=\linewidth]{figures/stats_barplot.png}
%         \caption{Rating distribution.}
%         \label{fig:fig_stat}
%     \end{subfigure}
%     \hfill
%     \begin{subfigure}[t]{0.48\linewidth}
%         \centering
%         \includegraphics[width=\linewidth]{figures/topic-new.png}
%         \caption{Top 10 topic distribution.}
%         \label{fig:fig_stat2}
%     \end{subfigure}
%     \caption{Rating and topic statistics of our dataset.}
%     \label{fig:fig_combined}
% \end{figure}
We collect 6,765 research papers from ICLR and NeurIPS (2023–2024), including both accepted and rejected submissions,
and filtered 5,687 usable data.
These papers cover diverse ML-related domains and topics.
% Topics and rating statistics are reported in Appendix~\ref{appendix:data}.
% 4,271 of the papers are used for training, with a separate set of 500 papers sampled for evaluation,
Each paper includes \emph{Abstracts, Methodology, and Experiment} sections\footnote{Paper content scraped from Semantic Scholar (\url{https://www.semanticscholar.org/product/api}) and ArXiv (\url{https://arxiv.org/help/api}) APIs, and then cleaned with regular expressions.}, supplemented with human reviews automatically obtained from OpenReview\footnote{\url{https://docs.openreview.net/reference/api-v2}.} for novelty, feasibility, effectiveness, and overall ratings.
Statistics of topics and rating distributions are reported in Appendix~\ref{appendix:data}.
Our data processing is released following the corresponding protocols.\footnote{\url{https://github.com/du-nlp-lab/LDC}}

% Paper content is 
% These papers and ratings are used to:
% 1. Derive ground-truth ideas for supervised fine-tuning.
% 2. Train reward models for the key dimensions.
% 3. Optimize idea generation using reinforcement learning with multi-dimension steering.
The dataset is split into the following subsets: 1) \emph{Supervised Fine-Tuning} split: 1,000 ICLR papers to derive the golden ideas and the most supporting paper for fine-tuning;
2) \emph{Reinforcement Learning} split:
    3,271 papers with detailed reviews to train reward models for novelty, feasibility, and effectiveness; and
3) \emph{Evaluation} split: 
    500 sampled papers for evaluation, including 30 randomly selected for manual expert review. 
    
 % 500 papers of the evaluation split for automatic evaluation, and a subset of 30 papers are selected for manual expert evaluation. We measure performance across three core metrics (details in Appendix):
% \begin{itemize}
%     \item \textbf{Novelty}: Evaluates how original the generated ideas are, compared to existing works.
%     \item \textbf{Feasibility}: Assesses the practical implementation and the likelihood that the idea can be executed within typical resource constraints.
%     \item \textbf{Effectiveness}: Measures the potential improvement or impact of the generated idea compared to baseline models.
% \end{itemize}

To ensure data reliability, we implement multi-stage quality control across automated extraction, retrieval, and filtering. 
We conduct a manual audit of 100 examples on topical match, plausibility, and completeness. Full details of the data processing and quality checks are provided in Appendix~\ref{appendix:quality}.
\vspace{-1mm}

\subsection{Evaluation Settings}
\vspace{-1mm}

% The evaluation is performed on two datasets: 
% the evaluation split (500 papers) and a manual subset (30 papers) using three metrics (details in Appendix):
% We split our evaluation into two types:
% \begin{enumerate}
%     \item \textbf{Automatic Evaluation}: For automatic evaluation, we evaluate novelty, feasibility, and effectiveness of the generated ideas with prompt-based method and GPT-4 as reviewing agent.
%     \item \textbf{Manual Evaluation}: For manual evaluation, we select 30 papers and have 15 domain experts assess the quality of the generated ideas. Details are described in Appendix~\ref{appendix:mannual}.
The evaluation is conducted following the settings in recent works~\cite{si2024llmsgeneratenovelresearch,baek2024researchagent}.
% on two datasets: the main evaluation split with a smaller subset for in-depth human evaluations.
We evaluate three key dimensions—\emph{novelty, feasibility, and effectiveness}—using both automatic and manual evaluation following the standard definition from OpenReview and human study~\cite{si2024llmsgeneratenovelresearch} as covered in Appendix \ref{appendix:def}.\\
\textbf{Automatic Evaluation.}
Following the recent trends in using LLMs to judge the quality of generated ideas~\cite{yang2024large, baek2024researchagent}, we use a prompt-based method with \textit{GPT-4} as the reviewing agent to score the generated ideas on all three metrics.
Different from their reference-free evaluations, we employ retrieval-augmented evaluation by fetching the latest related work from Semantic Scholar to ensure more faithful evaluations, especially for novelty.
We further validate the validity of this approach by measuring its correlation with human expert ratings.\\
\textbf{Manual Evaluation.}

% For manual evaluation, we randomly select a subset of 30 papers and recruit 15 domain experts from different institutes.
% All experts meet the reviewer criteria of leading venues such as NeurIPS, ACL, and EMNLP.
% Each expert is assigned only ideas within their own field of expertise, and each idea is scored independently by three experts on novelty, feasibility, and effectiveness---using the same OpenReview-derived rubrics as the automatic evaluation (Appendix~\ref{appendix:def})---together with an overall score and a mandatory written justification for every rating.
% As quality control, ratings are collected independently without discussion among annotators, and we quantify their reliability via inter-annotator agreement (Section~\ref{sec:human_eval}); the full recruiting criteria, annotation procedure, and example expert feedback are provided in Appendix~\ref{appendix:mannual}.
For manual evaluation, we randomly select a subset of 30 papers and have 15 domain experts across different institutes, recruited according to reviewer criteria adopted by leading conferences (e.g. NeurIPS, ACL, EMNLP) to independently assess ideas highly relevant to their field of expertise to assign a score for each criteria.
Each idea is rated by three experts, and they are also required to provide written justifications for their ratings.
Each dimension is scored following the same OpenReview-derived rubrics in Appendix~\ref{appendix:def}. All ratings are collected independently without discussion among annotators.
We also report inter-annotator agreement and manual feedback examples as in Appendix~\ref{appendix:mannual}, along with further details on the recruiting criteria and annotation process.

% \begin{tabular}{|c|c|c|}
% \hline
% \multirow{3}{*}{Group A} & x1 & y1 \\ 
%                          & x2 & y2 \\ 
%                          & x3 & y3 \\ \hline
% \end{tabular}

\begin{table*}[t]
\centering
\resizebox{\linewidth}{!}{
\begin{tabular}{ll|ccc|c}
\toprule
\textbf{Model} & & \textbf{Novelty(N)} & \textbf{Feasibility(F)} & \textbf{Effectiveness(E)} & \textbf{Overall} \\ 
\midrule
\multirow{2}{*}{\textbf{\textit{Baselines}}} & \textit{ResearchAgent~\cite{baek2024researchagent}} & 5.2 & 6.0 & 5.3 & 5.3\\
&\textit{MHABTO~\cite{su2025headsbetteroneimproved}} & 5.4 & 5.9	& 5.3 & 5.4\\
\midrule
\multirow{2}{*}{\textbf{SFT}}
 &\textit{T5} & 3.3 & 5.1 & 4.2 & 4.0 \\
&\textit{LLaMA-2-7B} & 4.5 & 5.6 & 5.2 & 5.1 \\
\midrule
\multirow{4}{*}{\textbf{\textit{SFT + RLHF}}}
& \textit{T5} & 3.8 & 5.3 & 4.8 & 4.5 \\ 
&\textit{Qwen-2.5-3B} &	5.0	& 5.7 & 5.2 & 5.3\\
& \textit{LLaMA-2-7B} & 5.3 & 6.0 & 5.5 & 5.4 \\ 
&\textit{LLaMA-3-8B} & 5.4 & 5.7 & 5.6 & 5.7\\
\midrule
\multirow{3}{*}{\makecell[l]{\textbf{\textit{SFT + RLHF}}\\\textit{\quad + Single Ctrl}}}
& \textit{LLaMA-3} + \textit{ Novelty} & 6.3$^{*}$ & 5.3 & 5.4 & 5.8\\
& \textit{LLaMA-3} + \textit{ Feasibility} & 5.1 & 6.8$^{*}$ & 5.3 & 5.3 \\ 
& \textit{LLaMA-3} + \textit{ Effectiveness} & 5.4 & 5.8 & 6.3$^{*}$ & 5.7 \\ 
\midrule
\multirow{3}{*}{\makecell[l]{\textbf{\textit{SFT + RLHF}}\\\textit{\quad + All Ctrls (Static)}}}
& \textit{Qwen-2.5-3B}	& 5.4	&6.0	&5.6	&5.6\\
& \textit{LLaMA-2-7B} & 5.4 & 5.9 & 5.5 & 5.5 \\ 
& \textit{LLaMA-3-8B}	&5.9	&6.3	&6.0	& 6.1\\
\midrule
\multirow{3}{*}{\makecell[l]{\textbf{\textit{SFT + RLHF}}\\\textit{\quad + All Ctrls (Dynamic)}}}
& \textit{Qwen-2.5-3B}	&5.6	&5.9	&5.8	&5.9$^{*}$\\
& \textit{LLaMA-2-7B} & 5.7 & 6.1 & 5.8 & 5.8 $^{*}$\\ 
& \textit{LLaMA-3-8B}	&6.2 &	6.4	&6.1	&6.3$^{*}$\\
\bottomrule
\end{tabular}}
\vspace{-2mm}

\caption{Experiment results with retrieval-augmented evaluation. \textit{Single Ctrl} denotes that only the mentioned controller is enabled. \textit{All Ctrls} activate all three controllers. \textit{Static} and \textit{Dynamic} denote different decoding strategies. * Significance checked against all baselines with p $<$ 0.05.}
\label{tab:main}
\vspace{-3mm}
\end{table*}

\vspace{-1.5mm}

\subsection{Main Experiments}
\paragraph{Baselines and Setups.}
We include comprehensive baselines and ablations to evaluate the effectiveness of different controls.
We compare against recent top-tier models, including \textit{ResearchAgent}~\cite{baek2024researchagent}, an agentic ideation with citation knowledge graph retrieval, and Many Heads Are Better Than One \textit{(MHABTO)}~\cite{su2025headsbetteroneimproved} with multi-agent simulation for ideation.
While AI Scientist~\cite{AIScientist} is relevant, its focus is not comparable. Comparability of related works are in Appendix~\ref{appendix:compare}, and a reference comparison with ChatGPT under the same evaluation protocol is reported in Appendix~\ref{appendix:chatgpt}.
We also include \textbf{\textit{SFT}} and \textbf{\textit{SFT + RLHF}} settings, trained with \textit{T5}, \textit{LLaMA-2-7B}, \textit{LLaMA-3-8B}, and \textit{Qwen-2.5-3B}:
\noindent \textit{SFT} is simply fine-tuned on 1,000 examples, while \textit{SFT + RLHF} is optimized with RL, but without dimensional controllers.
% \noindent \textbf{\textit{LLaMA2-SFT}} is LLaMA2-7B fine-tuned on 1,000 examples without reinforcement learning or control mechanisms.
% This baseline incorporates a more sophisticated retrieval pipeline to gather relevant literature and context for idea generation.\\
The RL split is used for both RL and dimensional controllers training.
% to optimize the model with PPO and multi-dimension reward augmentation.
% We incorporate the three distinct reward models for novelty, feasibility, and effectiveness, allowing for controllable generation combined with 3 control parameters, and experiment with different decoding strategies.
The three reward models (novelty, feasibility, effectiveness) enable controllable generation via tunable control parameters, and we experiment with static and dynamic decoding strategies.
% for ablation analysis.
The dimensional controllers $\mathbf{W}_{n/f/e}$ are trained with $\epsilon_{n/f/e}=1$, and the decoder RNN is with a hidden size of 256. The detailed implementation configurations of different stages are reported in Appendix~\ref{appendix:hyper}.

% For each setup:
% \begin{itemize}
%     \item \textbf{T5+T5}: In this configuration, both the reward models and the policy model were based on the T5 model. The reward models provided feedback on Novelty, Feasibility, and Effectiveness, while the T5 policy model generated and iteratively refined the ideas.
%     \item \textbf{LLaMA2-7b-hf}: In this configuration, we used the LLaMA2-7b-hf model for both the reward models and the policy model. Similar to the T5 setup, the reward models scored the generated ideas, which the LLaMA2-7b-hf policy model iterative improved.
% \end{itemize}

% We evaluated the agent's performance based on the feedback from the reward models, focusing on the balance between Novelty, Feasibility, and Effectiveness.
% % \paragraph{Metrics:}
% The generated ideas are evaluated with prompt-based reviewing agent and human expert across the three metrics: novelty, feasibility and effectiveness.
% In addition to individual scores, we also report an overall assessment score to give a holistic view of each model's performance.
% \vspace{-1.5mm}
\paragraph{Main Results.}
Table \ref{tab:main} summarizes the experimental results. While RLHF shows modest improvements for T5 in feasibility and effectiveness, the novelty remains limited.
LLaMA and Qwen achieve higher overall scores due to their larger capacity, but all benefit further from reinforcement learning and control strategy.
Adding targeted control to LLaMA3 with RLHF enables metric-specific
% \vspace{-1.5mm}
\begin{table}[t]
\centering
\resizebox{\linewidth}{!}{%
\begin{tabular}{l|ccc|c}
\toprule
\textbf{Model} & \textbf{N} & \textbf{F} & \textbf{E} & \textbf{Overall} \\
\midrule
% \textit{LLaMA2-SFT} & 4.31 & 5.59 & 4.83 & 4.59 \\
% \textit{LLaMA2-RLHF} & 4.94 & 6.18 & 5.20 & 5.33 \\ 
% \textit{LLaMA2-RLHF + Steers} & 5.49 & 6.36 & 5.11 & 5.51 \\ 
\textit{ResearchAgent}&	4.9 &	5.8	&5.1	&5.2\\
\midrule
\textit{LLaMA-2}\textit{-SFT} & 4.2 & 5.6 & 4.6 & 4.4 \\
% \textit{LLaMA-3}\textit{-SFT} & 4.5 & 5.7 & 5.0 & 5.2\\
\textit{LLaMA-2}\textit{-RLHF} & 4.9 & 6.0 & 5.1 & 5.3 \\ 
\textit{LLaMA-3}\textit{-RLHF} & 5.2 & 5.7 & 5.5 & 5.4\\
\textit{Qwen-2.5}\textit{-RLHF + Dynamic*}	&5.1	&5.9	&5.7	&5.5 \\
\textit{LLaMA-2}\textit{-RLHF + Dynamic*} & 5.3 & 6.2 & 5.2 & 5.6 \\ 
\textit{LLaMA-3}\textit{-RLHF + Dynamic*}	&5.7	&6.4	&5.4	&5.8 \\
\bottomrule
\end{tabular}}
\vspace{-1mm}
\caption{Human evaluation results. * denote dynamic decoding with all 3 controllers enabled.}
\vspace{-1mm}
\label{tab:human_eval}
\end{table}
\begin{table}[t]
\small
\centering
\resizebox{0.9\linewidth}{!}{%
\begin{tabular}{l|ccc|c}
\toprule
\textbf{Metrics} & \textbf{N} & \textbf{F} & \textbf{E} & \textbf{Overall} \\
\midrule
\textit{Pearson (r)}  & 0.982 & 0.948 & 0.716 & 0.871 \\
\textit{Spearman (p)} & 0.955 & 0.937 & 0.764 & 0.964 \\
\bottomrule
\end{tabular}}
\caption{Correlation coefficients (Pearson and Spearman) between human and reviewing agent scores.}
\vspace{-4mm}
\label{tab:correlation_metrics}
\end{table}
optimization and enhances its respective target dimension: Novelty control boosts creativity, with feasibility setting enhances practicality, and effectiveness improves impact.
Combining all controls, dynamic decoding outperforms the static approach across all metrics, balancing creativity, practicality, and impact effectively. Paired t-tests validate the significance. These highlight the importance of RL and dynamic control in optimizing model performance across complex requirements.
Notably, although ResearchAgent employs more advanced retrieval, our controllable models outperform it on all metrics, highlighting the effectiveness of controllable generation over complex retrieval alone.

\subsection{Human Evaluation Results}
\label{sec:human_eval}

\begin{table*}[t]
\small
    \centering
    \resizebox{\linewidth}{!}{

    \begin{tabular}{c|p{7.2cm}|c|c}
        \toprule
        \textbf{Model} & \textbf{Idea (Method part)} & \textbf{Novelty} / \textbf{Feasibility} / \textbf{Effectiveness} & \textbf{Overall} \\
        \midrule
        \textit{T5-SFT} & Proposing a reinforcement learning algorithm with stochastic agent interactions, focusing on decentralized learning in dynamic environments. The method avoids shared policies and uses predefined heuristics for adaptability. & 3.3 / 6.0 / 4.2 & 3.8 \\
        \midrule
        \textit{LLaMA-SFT} & Developing a reinforcement learning model that employs implicit environmental feedback for agent collaboration. The method eliminates the need for direct communication and uses fixed reward functions for learning. & 4.8 / 5.9 / 5.2 & 5.3 \\
        \midrule
        \textit{LLaMA-RLHF} & Introducing a reinforcement learning algorithm that combines stochastic interactions with an adaptive reward mechanism. This method enables efficient multi-agent collaboration in dynamic environments while ensuring scalability and practical feasibility. & 5.5 / 6.2 / 5.6 & 5.8 \\
        \midrule
          \textit{LLaMA-RLHF-Dynamic}
        % \makecell[c]{ \textit{LLaMA3-RLHF}\\\textit{\quad + All Ctrls (Dynamic)}}
    & Presenting a multi-agent reinforcement learning approach where agents utilize minimal communication protocols and enhanced environmental feedback. The method dynamically adjusts learning strategies to improve effectiveness in real-world applications. & 6.3 / 6.4 / 6.8 & 6.6 \\
        \bottomrule
    \end{tabular}}
            % \vspace{-2mm}

    \caption{Comparison of ideas (method part) and scores with all settings consistent with main experiments.}
        \label{tab:case_study}
        % \vspace{-3mm}
\end{table*}

% \begin{figure}[h]
%     \centering
%     \includegraphics[width=\linewidth]{figures/Human.png}
%     \caption{Human Evaluation Results}
%     \label{fig:human_eval}
%     \vspace{-3mm}
% \end{figure}

% We conducted a human evaluation by involving 10 domain experts to qualitatively assess 50 generated ideas. They rated the creativity, practicality, and clarity of ideas from the best-performing models. Evaluators were blind to the model source to avoid bias.

% \textbf{Manual Evaluation:}

The human evaluation is rigorously conducted according to the manual evaluation setting.
Domain experts validated the effectiveness of our framework of generated ideas, as in Table~\ref{tab:human_eval},
with human scores showing a strong correlation with the automatic evaluation scores.
The Correlation Coefficients computed with both Pearson and Spearman between human and reviewing agent scores are shown in Table~\ref{tab:correlation_metrics}.
Inter-annotator agreement reported in Appendix~\ref{appendix:mannual} is substantial on feasibility and effectiveness (with Fleiss' $\kappa=0.70$ and $0.65$), but moderate on novelty ($\kappa=0.41$).
We notice that in Section~\ref{rm_validation}, novelty is also where the reward models agree least with experts, suggesting that novelty is intrinsically harder to judge than the other two dimensions.

Experts also highlighted the trade-off between novelty and feasibility, that the fine-tuned model with novelty steering produced more creative, though sometimes less practical, ideas compared to the equal-weighted model.

% We measured inter-rater reliability using **Cohen’s Kappa**, achieving a high score of 0.82, indicating significant agreement. RLHF + Steer1 was ranked highest for novelty, while RLHF + Steers was rated the most practical for feasibility.

% \subsection{Significance Testing}

% \begin{table}[h!] \centering \caption{Correlation between human evaluation and LLM-based automatic scores.} \label{tab:human_eval} \begin{tabular}{|l|c|c|c|} \hline \textbf{Metric} & \textbf{Novelty (r)} & \textbf{Feasibility (r)} & \textbf{Effectiveness (r)} \ \hline Pearson (r) & 0.85 & 0.81 & 0.79 \ \hline Spearman (p) & 0.83 & 0.78 & 0.77 \ \hline \end{tabular} \end{table}

% \begin{table}[h]
% \small
% \centering
% \resizebox{0.9\linewidth}{!}{%
% \begin{tabular}{l|ccc|c}
% \toprule
% \textbf{Metrics} & \textbf{N} & \textbf{F} & \textbf{E} & \textbf{Overall} \\
% \midrule
% \textit{Pearson (r)}  & 0.982 & 0.948 & 0.716 & 0.871 \\
% \textit{Spearman (p)} & 0.955 & 0.937 & 0.764 & 0.964 \\
% \bottomrule
% \end{tabular}}
% \caption{Correlation coefficients (Pearson and Spearman) between human and reviewing agent scores.}
% \vspace{-2mm}
% \label{tab:correlation_metrics}
% \end{table}
% \vspace{-1.5mm}

\subsection{Reliability of Reward Models}
\label{rm_validation}
\begin{table}[t]
\small
\centering
\resizebox{0.72\linewidth}{!}{%
\begin{tabular}{l|ccc}
\toprule
\textbf{Metrics} & \textbf{N} & \textbf{F} & \textbf{E} \\
\midrule
\textit{Pearson (r)}  & 0.74 & 0.83 & 0.77 \\
\textit{Spearman (p)} & 0.71 & 0.80 & 0.74 \\
\bottomrule
\end{tabular}}
\vspace{-1mm}
\caption{Correlation coefficients between human expert scores and reward model predictions~($p<0.001$).}
\label{tab:rm_validation}
\end{table}
While some of the reward labels have no direct scores in human reviews, they can be inferred from the review and paper content: novelty by comparing the idea against the latest retrieved related work together with the review comments, and feasibility from the complexity of the method and the reported human labor, cost, computational resources, and time.
To examine whether the reward models trained on these LLM-derived labels capture expert judgement, we compare their predictions with expert scores on 90 held-out ideas, which cover all 30 papers in the manual evaluation set and are generated by three models (SFT, RLHF, and RLHF with Dynamic Control); each idea is rated independently by three experts following the manual evaluation setting.
As shown in Table~\ref{tab:rm_validation}, all three reward models correlate well with the mean expert scores, with feasibility being the highest and novelty the lowest.
In addition, as PPO and the dynamic decoder mostly rely on relative comparisons among candidate ideas rather than absolute scores,  these rank correlations suggest that the learned reward signals provide a reasonable training signal for our framework.

% \vspace{-5mm}

\begin{table}[t]
\small
\centering
% \resizebox{\linewidth}{!}{
\begin{tabular}{c|p{1.2cm}p{1.2cm}}
\toprule
\textbf{Novelty Weight} & \textbf{N} & \textbf{F} \\
\midrule
1.0 & 6.4 & 6.1 \\ 
2.0 & 6.7 & 5.8 \\ 
3.0 & 7.0 & 5.3 \\ 
4.0 & 7.3 & 4.9 \\ 
\bottomrule
\end{tabular}
\vspace{-1mm}
\caption{Novelty(N) and Feasibility(F) trade-off by increasing the novelty controller weight.}
\vspace{-3mm}
\label{tab:novelty_tradeoff}
\end{table}

\section{Analysis}
% \vspace{-1.25mm}
\subsection{Novelty and Feasibility Trade-off}

\citet{si2024llmsgeneratenovelresearch} find that increasing novelty will likely reduce the feasibility of an idea. To test this idea, we control the weight of the novelty steer in RLHF with novelty ctrl and observed its impact on both novelty and feasibility scores. The results are shown in Table~\ref{tab:novelty_tradeoff}.
As expected, increasing the novelty steer weight leads to higher novelty scores but lower feasibility scores. This demonstrates the trade-off between generating highly creative ideas and ensuring their practical feasibility.

\subsection{Decoding Strategy Motivation}
\begin{figure}[t]
    \centering
    \includegraphics[width=\linewidth]{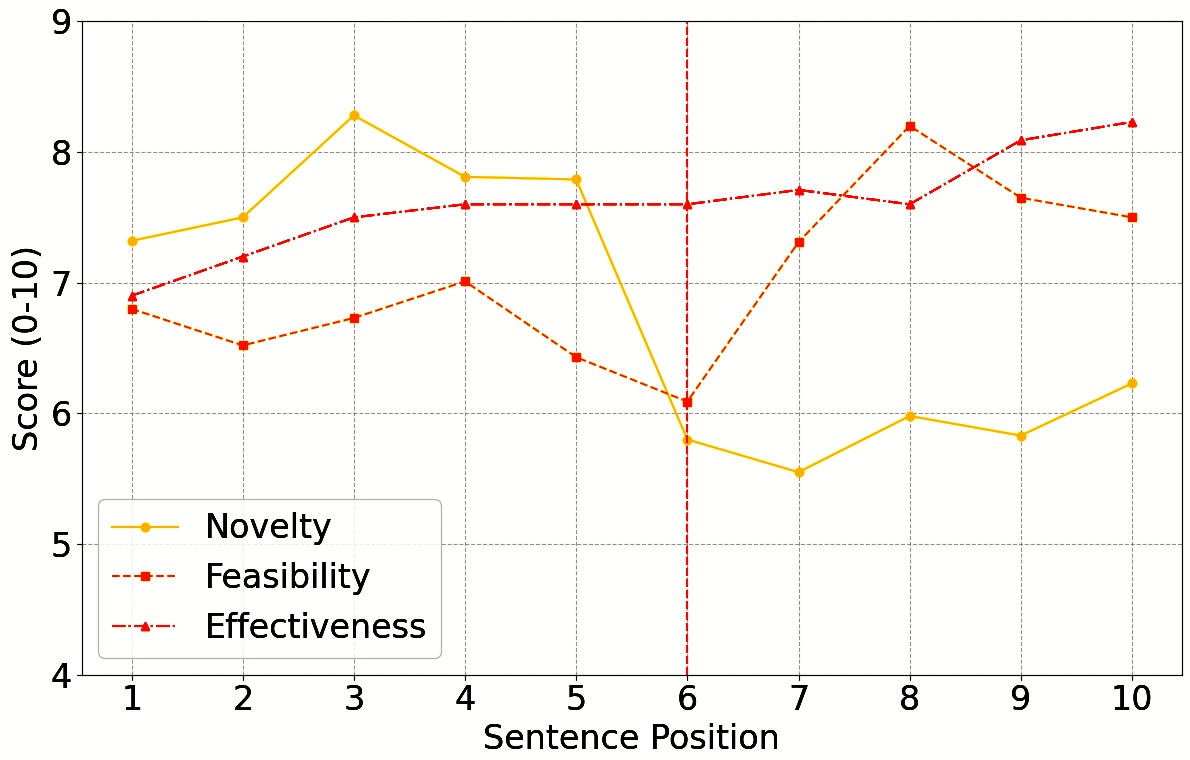}
    % \vspace{-5mm}
    \caption{Dimensional variation w.r.t. normalized sentence position (1-10 according to idea length).}
    % \vspace{-5mm}
    \label{fig:fig_nov_jump}
\end{figure}
Dynamic decoding adapts research ideation outputs to the varying demands of different parts of the idea, as shown in Figure~\ref{fig:fig_nov_jump}.
Note that all the sentences are normalized to 1-10 and put in the nearest integer bracket for better averaging. 
The observed novelty jump in the 6th sentence illustrates a shift in focus, aligning feasibility with the experiment plan while reducing the emphasis on novelty.
By dynamically adjusting decoding weights, this strategy ensures that the generated ideas are coherent, contextually aligned, and balanced across key dimensions. We further include a novelty and feasibility control analysis and scatter analysis in Appendix~\ref{appendix:control} and \ref{appendix:scatter}.

\subsection{Case Study}
% Idea comparison in Table \ref{tab:case_study} explores the increments of models from SFT to advanced configurations with adaptive control mechanisms. Baseline models like T5-SFT and LLaMA-SFT demonstrate moderate feasibility but fall short in balancing novelty and effectiveness. With RL fine-tuning, LLaMA2-RLHF clearly improves across all metrics by adding reward mechanisms, enhancing multi-agent collaboration in dynamic settings. The integration of steers further optimizes performance, with LLaMA-RLHF with control achieving the highest overall score through dynamic adjustments that balance creativity, practicality, and impact. This progression highlights the transformative potential of RL and dynamic control strategies in generating innovative, feasible, and impactful ideas.
Table \ref{tab:case_study} compares the evolution of ideas generated by models, progressing from SFT to advanced configurations with dynamic control. Baseline models with SFT exhibit moderate feasibility but struggle to achieve a balance between novelty and effectiveness, highlighting their limitations in fostering creative yet practical solutions. RLHF demonstrates clear improvements across all metrics, leveraging reward mechanisms to enhance collaboration of fine-grained dimensions. Adding dynamic control further elevates performance, achieving the highest overall. This progression underscores the potential of RL fine-tuning combined with context-aware dynamic control for innovative, practical, and highly effective idea generation. 
\vspace{-1mm}

% \vspace{-2mm}

\section{Conclusion}
\vspace{-2mm}
We present a novel framework with LLM for research idea generation that optimizes and dynamically balances key dimensions—novelty, feasibility, and effectiveness—through a two-stage process combining supervised fine-tuning and controllable reinforcement learning.
By leveraging multi-dimension reward models and integrating the dimensional controller with sentence-level dynamic decoding, our approach effectively navigates the improvement and the inherent trade-offs among these metrics, ensuring context-aware and high-quality idea generation.
Comprehensive evaluations, including human studies, highlight the robustness and effectiveness of our method, giving a path for more advanced and controllable systems in automated research idea generation.

% We investigate how 

\section*{Limitations}
Firstly, although the research ideas generated are of good quality, citation prediction could be explored as another way to judge their quality, which also makes it easier for human researchers to select promising ideas. Secondly, the interpretability of learned adjustments of dimension controllers is still a remaining open question for future exploration. Thirdly, some of the reward labels are inferred from the paper and review content through prompting. Although they correlate well with expert judgement, they may still inherit biases from the underlying reviews and the prompting model.

% \section*{Acknowledgements}

\bibliography{custom}

@inproceedings{valuefunc,
  author       = {John Schulman and
                  Philipp Moritz and
                  Sergey Levine and
                  Michael I. Jordan and
                  Pieter Abbeel},
  title        = {High-Dimensional Continuous Control Using Generalized Advantage Estimation},
  booktitle    = {{ICLR}},
  year         = {2016},
  url          = {http://arxiv.org/abs/1506.02438},
  bibsource    = {dblp computer science bibliography, https://dblp.org}
}

@article{yang2024large,
  title={Large Language Models Meet Text-Centric Multimodal Sentiment Analysis: A Survey},
  author={Yang, Hao and Zhao, Yanyan and Wu, Yang and Wang, Shilong and Zheng, Tian and Zhang, Hongbo and Che, Wanxiang and Qin, Bing},
  journal={arXiv preprint arXiv:2406.08068},
  year={2024}
}

@inproceedings{han2024word,
  title={Word Embeddings Are Steers for Language Models},
  author={Han, Chi and Xu, Jialiang and Li, Manling and Fung, Yi and Sun, Chenkai and Jiang, Nan and Abdelzaher, Tarek and Ji, Heng},
  booktitle={Proceedings of the 62nd Annual Meeting of the Association for Computational Linguistics (Volume 1: Long Papers)},
  pages={16410--16430},
  year={2024},
  organization={Association for Computational Linguistics}
}

@article{bornstein2024hypothesiscraft,
  title={HypothesisCraft: Towards Automated Hypothesis Generation and Refinement Using LLMs},
  author={Bornstein, Michael and Singh, Rahul},
  journal={Journal of AI Research},
  year={2024}
}

@inproceedings{beltagy2019scibert,
  title={SciBERT: A pretrained language model for scientific text},
  author={Beltagy, Iz and Lo, Kyle and Cohan, Arman},
  booktitle={Proceedings of the 2019 Conference on Empirical Methods in Natural Language Processing (EMNLP)},
  pages={3615--3620},
  year={2019}
}

@article{lee2020biobert,
  title={BioBERT: a pre-trained biomedical language representation model for biomedical text mining},
  author={Lee, Jinhyuk and Yoon, Wonjin and Kim, Sungdong and Kim, Donghyeon and Kim, Sunkyu and Kim, Chan and Kang, Jaewoo},
  journal={Bioinformatics},
  volume={36},
  number={4},
  pages={1234--1240},
  year={2020},
  publisher={Oxford University Press}
}

@article{raghu2020survey,
  title={A survey of deep learning for scientific discovery},
  author={Raghu, Maithra and Schmidt, Eric},
  journal={arXiv preprint arXiv:2003.11755},
  year={2020}
}

@misc{huang2024mlagentbench,
      title={MLAgentBench: Evaluating Language Agents on Machine Learning Experimentation}, 
      author={Qian Huang and Jian Vora and Percy Liang and Jure Leskovec},
      year={2024},
      eprint={2310.03302},
      archivePrefix={arXiv},
      primaryClass={cs.LG},
      url={https://arxiv.org/abs/2310.03302}, 
}

@article{GPT-4,
  author       = {OpenAI},
  title        = {{GPT-4} Technical Report},
  journal      = {arXiv preprint arXiv:2303.08774},
  year         = {2023},
  url          = {https://doi.org/10.48550/arXiv.2303.08774},
  doi          = {10.48550/ARXIV.2303.08774},
}

@article{baek2024researchagent,
  title={ResearchAgent: Iterative Research Idea Generation over Scientific Literature with Large Language Models},
  author={Baek, Jinheon and Jauhar, Sujay Kumar and Cucerzan, Silviu and Hwang, Sung Ju},
  journal={NAACL 2025},
  year={2025}
}

@inproceedings{Wang2023SciMONSI,
  title={{SciMON: Scientific Inspiration Machines Optimized for Novelty}},
  author={Qingyun Wang and Doug Downey and Heng Ji and Tom Hope},
  year={2024},
  booktitle={ACL}
}

@article{Yang2023LargeLM,
  title={{Large Language Models for Automated Open-domain Scientific Hypotheses Discovery}},
  author={Zonglin Yang and Xinya Du and Junxian Li and Jie Zheng and Soujanya Poria and E. Cambria},
  journal={ACL Findings},
  year={2024}
}

@article{Tian2024SciCodeAR,
  title={{SciCode: A Research Coding Benchmark Curated by Scientists}},
  author={Minyang Tian and Luyu Gao and Shizhuo Dylan Zhang and Xinan Chen and Cunwei Fan and Xuefei Guo and Roland Haas and Pan Ji and Kittithat Krongchon and Yao Li and Shengyan Liu and Di Luo and Yutao Ma and Hao Tong and Kha Trinh and Chenyu Tian and Zihan Wang and Bohao Wu and Yanyu Xiong and Shengzhu Yin and Min Zhu and Kilian Lieret and Yanxin Lu and Genglin Liu and Yufeng Du and Tianhua Tao and Ofir Press and Jamie Callan and E. A. Huerta and Hao Peng},
  year={2024},
  journal={ArXiv},
  volume={abs/2407.13168}
}

@article{AIScientist,
  title={{The AI Scientist: Towards Fully Automated Open-Ended Scientific Discovery
}},
  author={Chris Lu and Cong Lu and Robert Tjarko Lange and Jakob Foerster and Jeff Clune and David Ha},
  journal={ArXiv},
  year={2024},
  volume={abs/2408.06292},
}

@article{Li2024MLRCopilotAM,
  title={{MLR-Copilot: Autonomous Machine Learning Research based on Large Language Models Agents}},
  author={Ruochen Li and Teerth Patel and Qingyun Wang and Xinya Du},
journal={ArXiv},  
  year={2024},
  volume={abs/2408.14033}
}

@article{christiano2017deep,
  title={Deep reinforcement learning from human preferences},
  author={Christiano, Paul F and Leike, Jan and Brown, Tom B and Martic, Miljan and Legg, Shane and Amodei, Dario},
  journal={Advances in Neural Information Processing Systems},
  volume={30},
  year={2017}
}

@article{stiennon2020learning,
  title={Learning to summarize with human feedback},
  author={Stiennon, Nisan and Ouyang, Long and Wu, Jeffrey and Ziegler, Daniel M and Lowe, Ryan and Voss, Chelsea and Radford, Alec and Amodei, Dario and Christiano, Paul F},
  journal={Advances in Neural Information Processing Systems},
  volume={33},
  pages={3008--3021},
  year={2020}
}

@article{ouyang2022training,
  title={Training language models to follow instructions with human feedback},
  author={Ouyang, Long and Wu, Jeffrey and Jiang, Xu and Almeida, Diogo and Wainwright, Carroll L and Mishkin, Pamela and Zhang, Chong and Agarwal, Sandhini and Slama, Katarina and Ray, Alex and others},
  journal={arXiv preprint arXiv:2203.02155},
  year={2022}
}

@article{ziegler2019fine,
  title={Fine-tuning language models from human preferences},
  author={Ziegler, Daniel M and Stiennon, Nisan and Wu, Jeffrey and Brown, Tom and Radford, Alec and Amodei, Dario and Christiano, Paul F},
  journal={arXiv preprint arXiv:1909.08593},
  year={2019}
}

@article{nakano2021webgpt,
  title={WebGPT: Browser-assisted question-answering with human feedback},
  author={Nakano, Reiichiro and Hilton, Jacob and Balaji, Suchir and Wu, Jeff and Ouyang, Long and Kim, Christina and Jones, Matthew and Saunders, William and Hesse, Chris and Elhage, Nelson and others},
  journal={arXiv preprint arXiv:2112.09332},
  year={2021}
}

@article{glaese2022improving,
  title={Improving alignment of dialogue agents via targeted human judgements},
  author={Glaese, Amal and McAleese, Natasha and Trager, Samuel and Askell, Amanda and Bai, Yuntao and Chen, Andy and Drain, Dawn and Elhage, Nelson and Ganguli, Deep and Godwin, Jonathan and others},
  journal={arXiv preprint arXiv:2209.14375},
  year={2022}
}

@article{uesato2022helpful,
  title={Fine-grained reward models for helpful and harmless assistants},
  author={Uesato, Jonathan and Reichert, David and Mellor, Joe and Tils, David and Huang, Po-Sen and Cai, Trevor and Aravind, Arvind Neelakantan and Desai, Sankalp and Moats, Isaac and Glaese, Amal and others},
  journal={arXiv preprint arXiv:2209.11895},
  year={2022}
}

@article{hope2021scisight,
  title={SciSight: Combining faceted navigation and research group detection for COVID-19 exploratory scientific search},
  author={Hope, Tom and Portnoff, Rebecca and Briscoe, Erica and Krishnan, Rahul and Murphy, Kevin and Cohen, William W and Anandan, Padmanabhan and Yates, Andrew and Benton, Adrian and Daum{\'e} III, Hal and others},
  journal={arXiv preprint arXiv:2005.12683},
  year={2021}
}

@article{sherstinsky2020fundamentals,
  title={Fundamentals of recurrent neural network (RNN) and long short-term memory (LSTM) network},
  author={Sherstinsky, Alex},
  journal={Physica D: Nonlinear Phenomena},
  volume={404},
  pages={132306},
  year={2020},
  publisher={Elsevier}
}

@article{brown2020language,
  title={Language models are few-shot learners},
  author={Brown, Tom and Mann, Benjamin and Ryder, Nick and Subbiah, Melanie and Kaplan, Jared D and Dhariwal, Prafulla and Neelakantan, Arvind and Shyam, Pranav and Sastry, Girish and Askell, Amanda and others},
  journal={Advances in neural information processing systems},
  volume={33},
  pages={1877--1901},
  year={2020}
}

@article{zhong2023goal,
  title={Goal driven discovery of distributional differences via language descriptions},
  author={Zhong, Ruiqi and Zhang, Peter and Li, Steve and Ahn, Jinwoo and Klein, Dan and Steinhardt, Jacob},
  journal={Advances in Neural Information Processing Systems},
  volume={36},
  pages={40204--40237},
  year={2023}
}

@article{qi2023large,
  title={Large language models are zero shot hypothesis proposers},
  author={Qi, Biqing and Zhang, Kaiyan and Li, Haoxiang and Tian, Kai and Zeng, Sihang and Chen, Zhang-Ren and Zhou, Bowen},
booktitle={Proceedings of Instruction Tuning and Instruction Following at NeurIPS 2023},
  year={2023},
url={https://arxiv.org/pdf/2311.05965}
}

@article{DBLP:journals/corr/SchulmanWDRK17,
  author       = {John Schulman and
                  Filip Wolski and
                  Prafulla Dhariwal and
                  Alec Radford and
                  Oleg Klimov},
  title        = {Proximal Policy Optimization Algorithms},
  journal      = {CoRR},
  volume       = {abs/1707.06347},
  year         = {2017},
  url          = {http://arxiv.org/abs/1707.06347},
  eprinttype    = {arXiv},
  eprint       = {1707.06347},
  bibsource    = {dblp computer science bibliography, https://dblp.org}
}

@article{DBLP:journals/corr/abs-2406-10305,
  author       = {Jie Chen and
                  Xintian Han and
                  Yu Ma and
                  Xun Zhou and
                  Liang Xiang},
  title        = {Unlock the Correlation between Supervised Fine-Tuning and Reinforcement
                  Learning in Training Code Large Language Models},
  journal      = {CoRR},
  volume       = {abs/2406.10305},
  year         = {2024},
  url          = {https://doi.org/10.48550/arXiv.2406.10305},
  doi          = {10.48550/ARXIV.2406.10305},
  eprinttype    = {arXiv},
  eprint       = {2406.10305},
  bibsource    = {dblp computer science bibliography, https://dblp.org}
}

@misc{jing2024fgaifaligninglargevisionlanguage,
      title={FGAIF: Aligning Large Vision-Language Models with Fine-grained AI Feedback}, 
      author={Liqiang Jing and Xinya Du},
      year={2024},
      eprint={2404.05046},
      archivePrefix={arXiv},
      primaryClass={cs.CV},
      url={https://arxiv.org/abs/2404.05046}
}

@misc{su2025headsbetteroneimproved,
      title={Many Heads Are Better Than One: Improved Scientific Idea Generation by A LLM-Based Multi-Agent System}, 
      author={Haoyang Su and Renqi Chen and Shixiang Tang and Zhenfei Yin and Xinzhe Zheng and Jinzhe Li and Biqing Qi and Qi Wu and Hui Li and Wanli Ouyang and Philip Torr and Bowen Zhou and Nanqing Dong},
      year={2025},
      eprint={2410.09403},
      archivePrefix={arXiv},
      primaryClass={cs.AI},
      url={https://arxiv.org/abs/2410.09403}, 
}

@misc{si2024llmsgeneratenovelresearch,
      title={Can LLMs Generate Novel Research Ideas? A Large-Scale Human Study with 100+ NLP Researchers}, 
      author={Chenglei Si and Diyi Yang and Tatsunori Hashimoto},
      year={2024},
      eprint={2409.04109},
      archivePrefix={arXiv},
      primaryClass={cs.CL},
      url={https://arxiv.org/abs/2409.04109}, 
}

@inproceedings{DBLP:conf/nips/WuHSDSASOH23,
  author       = {Zeqiu Wu and
                  Yushi Hu and
                  Weijia Shi and
                  Nouha Dziri and
                  Alane Suhr and
                  Prithviraj Ammanabrolu and
                  Noah A. Smith and
                  Mari Ostendorf and
                  Hannaneh Hajishirzi},
  title        = {Fine-Grained Human Feedback Gives Better Rewards for Language Model
                  Training},
  booktitle    = {Advances in Neural Information Processing Systems 36: Annual Conference
                  on Neural Information Processing Systems},
  year         = {2023}
}

@inproceedings{wang2024scimonscientificinspirationmachines,
    title     = {SciMON: Scientific Inspiration Machines Optimized for Novelty},
    author    = {Wang, Qingyun and Downey, Doug and Ji, Heng and Hope, Tom},
    year      = {2024},
    title = {Proceedings of the 62nd Annual Meeting of the Association for Computational Linguistics (ACL2024)}
}
\bibliographystyle{acl_natbib}

% \newpage
\newpage
% \FloatBarrier

\newpage
\onecolumn
\setlength{\textwidth}{\dimexpr\pdfpagewidth-\oddsidemargin-\evensidemargin-2in}
% \twocolumn

% \section*{Appendix}
\clearpage
\appendix
% \section*{Appendix}

\section{PPO and Detailed Algorithm for Multi-dimension reward augmented RL}
\label{appendix:ppo}
\begin{algorithm}[h]
\caption{Multi-dimension reward augmented Reninformace Learning}
\label{alg:RLHF}
\textbf{Input:} Initial policy model $\mathcal{M}_{\theta_{init}}$; initial value model $V_{\psi_{init}}$; $3$ well-trained reward models $\mathcal{R}_{n/f/e}$; 
task prompts $\mathcal{D}$; hyperparameters $\gamma $, $ \lambda $, $\epsilon$ \\
\textbf{Output:} Updated policy models $\mathcal{M}_{\theta_{n/f/e}}$.

\begin{algorithmic}[H]
    \State Initialize policy model $\mathcal{M}_{\theta_{n/f/e}} \gets \mathcal{M}_{\theta_{init}}$, value model $V_\psi^{n/f/e} \gets V_{\psi_{init}}$
    \For{step $= 1, \dots, M$}
        \State Sample a batch $\mathcal{D}_b$ from $\mathcal{D}$
        \State Sample output sequence $y^n_n \sim \mathcal{M}_{\theta_n} (\cdot \mid x^n)$, $y^n_f \sim \mathcal{M}_{\theta_f}(\cdot \mid x^n)$, $y^n_e \sim \mathcal{M}_{\theta_e} (\cdot \mid x^n)$ for each prompt $x^n \in \mathcal{D}_b$
        \State Compute rewards $\{r_t^{n/f/e}\}_{t=1}^{|y^n|}$ for each sampled output $y^n_n, y^n_f, y^n_e$ by running $\mathcal{R}^{o/a/r}$ \hfill
        \State Compute advantages $\{A_t^{o/a/r}\}_{t=1}^{|y^n|}$ and value targets $\{V_{\text{targ}}^{o/a/r}(s_t)\}_{t=1}^{|y^n|}$ for each $y^n_n, y^n_f, y^n_e$ with $V_\psi$
        \For{PPO iteration $= 1, \dots, \mu$}
            \State Update the policy model by maximizing the PPO clipped surrogate objective for $\mathcal{M}_{\theta_{n/f/e}}$:
            \begin{equation*}
            \begin{aligned}
                &\theta \gets \arg\max_\theta \frac{1}{|\mathcal{D}_b|} \sum_{n=1}^{|\mathcal{D}_b|} \frac{1}{|y^n|} \sum_{t=1}^{|y^n|} \min( \\ &\frac{\mathcal{M}_\theta(a_t \mid s_t)}{\mathcal{M}_{\theta_{\text{old}}}(a_t \mid s_t)} A_t, \text{clip}(v_t, 1-\epsilon, 1+\epsilon)A_t )
            \end{aligned}
            \end{equation*}
        \EndFor
        \State Update the value model $\psi_{n/f/e}$ by minimizing a square-error objective:
        \begin{equation*}
        \begin{aligned}
            &\psi \gets \arg\min_\psi \frac{1}{|\mathcal{D}_b|} \sum_{n=1}^{|\mathcal{D}_b|} \frac{1}{|y^n|} \sum_{t=1}^{|y^n|} ( V_\psi(s_t) - \\ & V_{\operatorname{targ}}(s_t) )^2
             \end{aligned}
        \end{equation*}
    \EndFor
\end{algorithmic}
\end{algorithm}
To optimize our idea proposer, we utilize Proximal Policy Optimization (PPO), an actor-critic RL algorithm widely used in previous RLHF works.
PPO enables the proposer (i.e. the policy model) to be refined against multiple reward models that simulate human feedback, ensuring high-quality idea generation.
In PPO, the value model $V_\psi(s_t)$ estimates the expected cumulative reward for a given state $s_t$, providing a baseline for the advantage function.
% to stabilize training.
The proposer is optimized with a PPO clipped surrogate training objective.
The advantage $A_t$ at timestep $t$ is estimated by a generalized advantage estimation function \citep{valuefunc}: $A_t = \sum_{t'=t}^{T} (\gamma\lambda)^{t'-t} (r_{t'} + \gamma V_\psi(s_{t'+1}) - V_\psi(s_{t'}))$,
with $\gamma$ as a hyperparameter and $\lambda$ as the discounting factor for rewards. $r_t$ is the reward assigned to $a_t$, which in our case is acquired using multiple learned reward models.
The value model $V_\psi(s_t)$ is optimized with an expected squared-error loss with the value target as
$V_{\text{targ}}(s_t) = \sum_{t'=t}^{T-1} \gamma^{t'-t} r_{t'} + \gamma^{T-t} V_{\psi_{\text{old}}}(s_T)$,
where $V_{\psi_{\text{old}}}$ is the lagging value model.
Finally, PPO is trained to optimize both the proposer ($\mathcal{M}_\theta$) and value ($V_\psi$) models with their respective objectives.
No reward model is being optimized during PPO training. See Algorithm \ref{alg:RLHF} for more details.

\newpage

\section{Data Statistics}
\label{appendix:data}
\begin{figure}[h]
    \centering
    \begin{subfigure}[t]{0.5\linewidth}
        \centering
        \includegraphics[width=\linewidth]{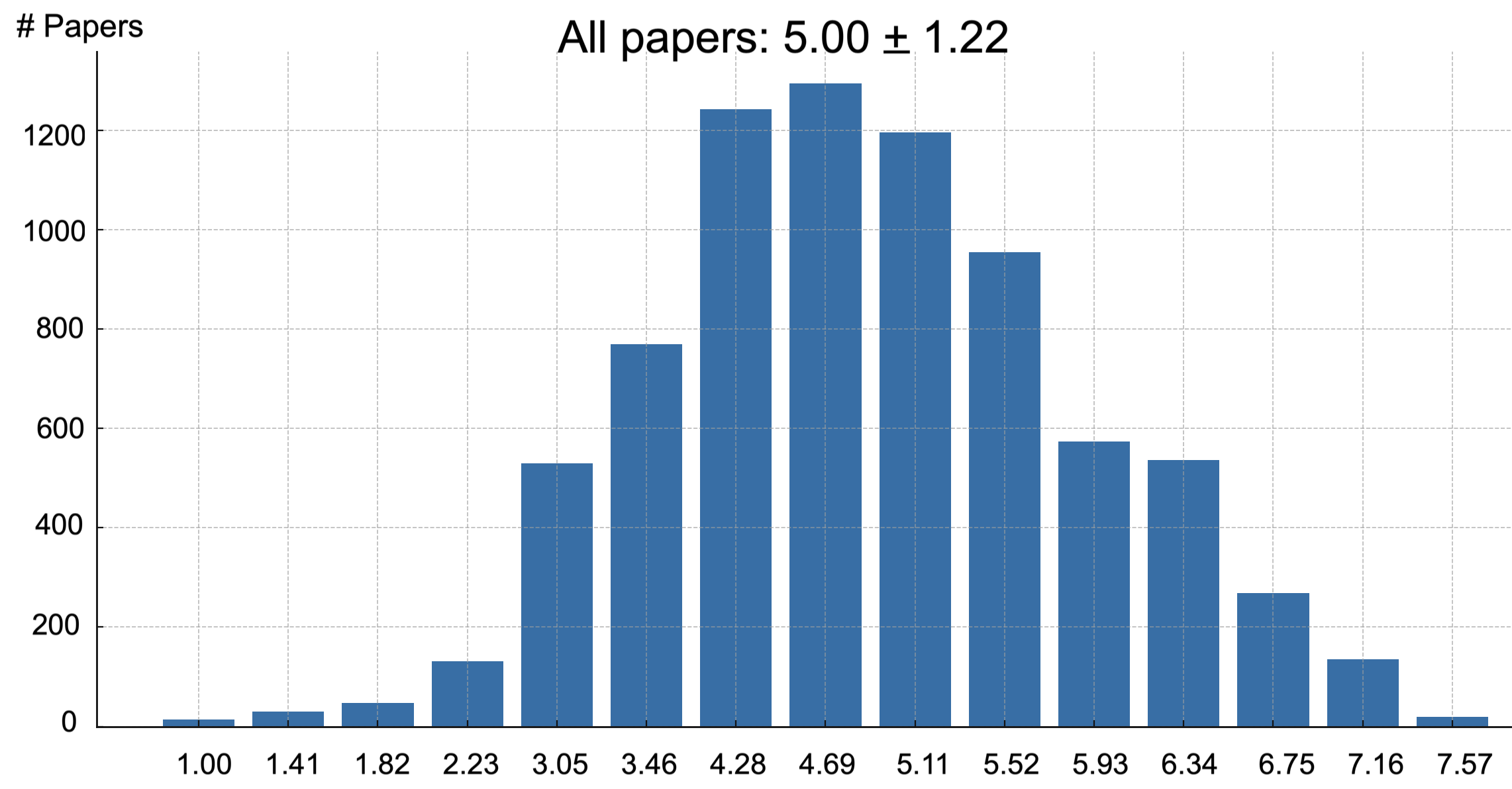}
        \caption{Rating distribution.}
        \label{fig:fig_stat}
    \end{subfigure}
    \vspace{5em}
    \begin{subfigure}[t]{0.6\linewidth}
        \centering
        \includegraphics[width=\linewidth]{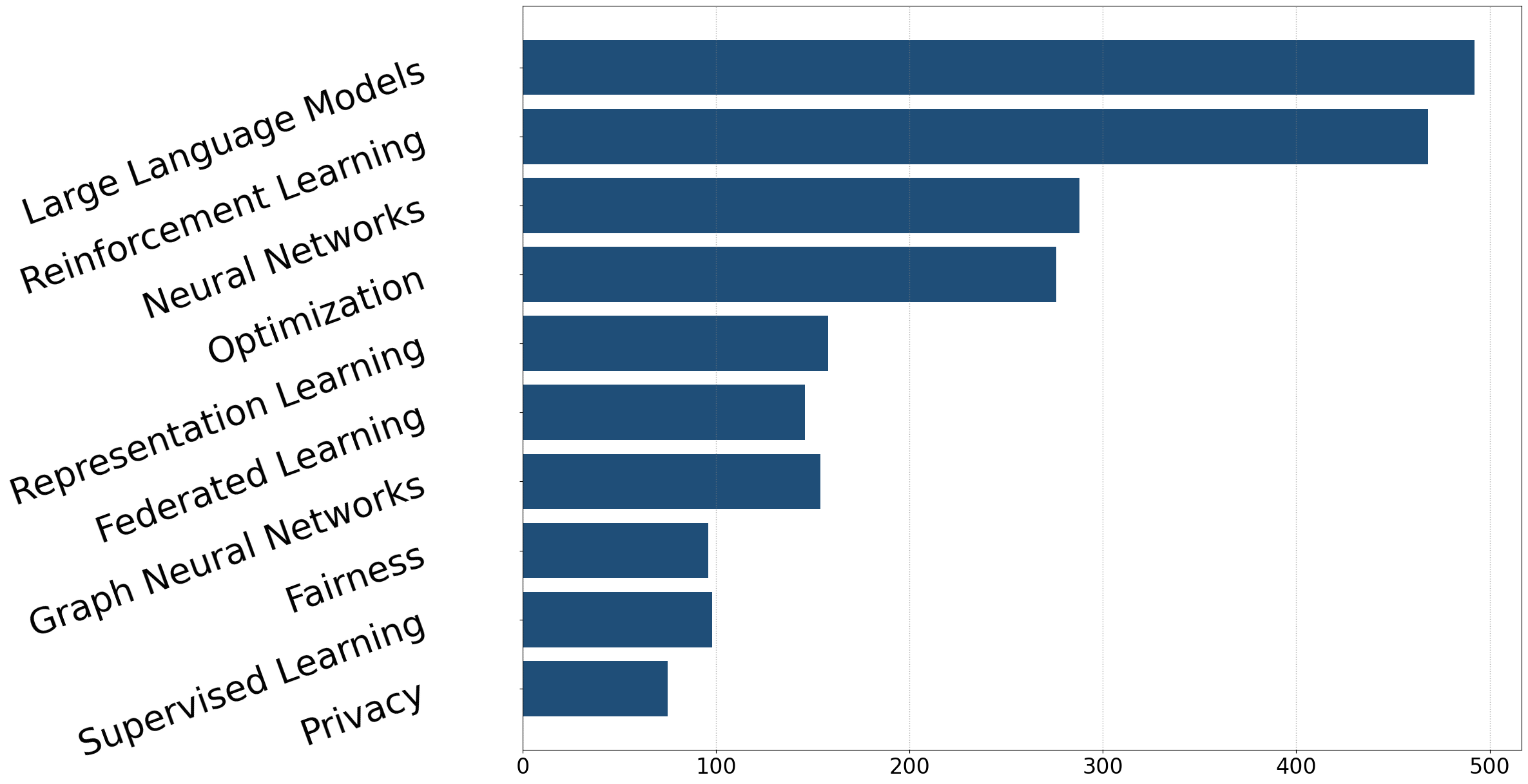}
        \caption{Top 10 topic distribution.}
        \label{fig:fig_stat2}
    \end{subfigure}
    \caption{Rating and topic statistics of our dataset.}
    \label{fig:fig_combined}
\end{figure}

Figures \ref{fig:fig_combined} provide an overview of the dataset distribution and top keywords.

\section{Quality Control of SFT Data}
\label{appendix:quality}
To ensure the quality and relevance of the supervised fine-tuning (SFT) data, we implement a multi-stage quality control in terms of the following aspects:
\begin{itemize}
    \item \textbf{Source and Extraction}: Ideas are not freely generated, but are extracted from ICLR/NeurIPS 2023–2024 papers. We prompt LLaMA3 to extract the central research idea from each paper's abstract and introduction.
    \item \textbf{Supporting Paper Selection}: The associated main supporting paper is identified through a retrieval process that integrates citation graph statistics with LLaMA3-based prompting.
    \item \textbf{Screening Filters}: We apply automated filters to remove incoherent, or incomplete off-topic samples.
    \item \textbf{RL-based Optimization}: We rely on RL-based fine-tuning for subsequent optimization, using data that has been evaluated by human experts to ensure high performance (see Section~\ref{sec:rl}).
    \item \textbf{Human Feedback in Reward Modeling}: The OpenReview data used for reward modeling is manually scored to reflect human judgments.
\end{itemize}
\vspace{2mm}
\textbf{Manual Audit of SFT Samples:}\\
\vspace{1mm}
To further evaluate data quality, we (authors) conducted a manual audit of 100 randomly sampled SFT examples. Three criteria were evaluated on a scale from 0 to 10:
\begin{itemize}
    \item \textbf{Paper-Idea Topical Match:} How well the extracted idea matches the main topic of the paper.
    \item \textbf{Plausibility of Idea:} Whether the idea is realistic and logically follows from the paper context.
    \item \textbf{Completeness:} Whether the extracted idea is sufficiently complete and self-contained.
\end{itemize}
These quality control procedures and manual audit results in Table~\ref{tab:std} demonstrate that our SFT dataset is generally of high quality and well-suited for model training.
\vspace{-2mm}
\begin{table}[h]
\small
\centering
\begin{tabular}{lcc}
\toprule
Criterion & Mean Score (0--10) & Std. Dev. \\
\midrule
Paper-Idea Topical Match & 8.1 & 1.2 \\
Plausibility of Idea     & 7.8 & 1.4 \\
Completeness             & 7.5 & 1.6 \\
\bottomrule
\end{tabular}
\caption{Manual audit of 100 randomly selected SFT samples: mean scores and standard deviations.}
\label{tab:std}
\end{table}

\section{Manual Evaluation Details}
\label{appendix:mannual}
For manual evaluation, we randomly select 30 papers and have 15 domain experts from different institutes (including several faculty members) assess the quality of the generated ideas for each model (SFT, RLHF, and RLHF with Dynamic Controls), with each idea independently annotated by three experts.
% \textcolor{blue}{
To ensure the rigor and authority of human evaluation, all annotators meet widely accepted reviewer criteria used by leading conferences such as NeurIPS, ACL, and EMNLP. Specifically, our experts satisfy a combination of the following requirements:
\begin{itemize}
    \item Hold a PhD or are authors of multiple peer-reviewed publications in relevant fields;
    \item Have at least two first-author publications in major conferences or journals (e.g., NeurIPS, ACL, EMNLP, ICML, ICLR, etc.) within the past five years;
    \item Have served as a reviewer in these conferences or journals, or have demonstrated substantial research expertise via citation record and research experience.
\end{itemize}
For each evaluation, annotators are required to provide a written justification for their ratings. On average, each evaluation took approximately three minutes to complete. Each expert provides human scores for novelty, feasibility, and effectiveness, which are then compared with those generated by our automatic reviewing agent to measure the alignment between human judgment and the agent's evaluations.\\\\
Additionally, we conduct inter-annotator agreement for all evaluation criteria to quantify the consistency among experts. The average Fleiss' kappa scores across all criteria are \textit{novelty: 0.41}, \textit{feasibility: 0.70}, and \textit{effectiveness: 0.65}, reflecting good inter-annotator agreement.
We also collect and analyze written feedback from annotators to better understand the qualitative aspects of novelty, feasibility, and effectiveness.\\
A representative example of human evaluation is below:
\begin{itemize}
    \item \textbf{Idea:} \textit{We tackle multimodal mental health assistance (text + voice tone + facial expressions). We introduce adaptive fine-tuning with emotion and sentiment feedback for state tracking; and incorporate trust and transparency feedback drawing insights from explainable AI. Experimental plan with setup, dataset, baselines, metrics, ablation, and expected results...}
    \item \textbf{Scores:} Novelty = 7, Feasibility = 6, Effectiveness = 8, Overall = 7
    \item \textbf{Feedback:} `` This idea provides a novel multimodal setting to propose adaptive fine-tuning. It focuses on fine-grained aspects over single-score feedback. The experiment design is solid and appears feasible. Data collection may pose minor challenges.''
\end{itemize}

\section{Hyper-parameters and Implementation Details}
\label{appendix:hyper}
Table~\ref{tab:hyper} lists the training configuration for the PPO stage, the dimensional controllers, and the sentence-level decoder.
\begin{table}[h]
\small
\centering
\begin{tabular}{ll@{\qquad}ll@{\qquad}ll}
\toprule
\multicolumn{2}{l}{\textbf{PPO}} & \multicolumn{2}{l}{\textbf{Dimensional controllers}} & \multicolumn{2}{l}{\textbf{Sentence-level decoder (RNN)}}\\
\cmidrule(r){1-2}\cmidrule(r){3-4}\cmidrule(r){5-6}
Learning rate & 1e-6 & Initialization ($\mathbf{W}_{n/f/e}$) & Xavier uniform & Hidden size & 256 \\
Clipping $\varepsilon$ & 0.2 & $\epsilon_{n/f/e}$ during training & 1.0 & $\epsilon_{\max}$ & 5.0 \\
Discount $\gamma$ & 1.0 & & & & \\
GAE $\lambda$ & 0.95 & & & & \\
PPO epochs & 4 & & & & \\
\bottomrule
\end{tabular}
\caption{Hyper-parameter configuration for RL training, dimensional controllers, and the sentence-level decoder.}
\label{tab:hyper}
\end{table}

\section{Why Train Three Controllers Independently?}
\label{appendix:steer_theory}
Following the notations in the method section, we use $\theta$ to denote the model parameters.
The independent-controller approach trains the model separately for each reward function:
\begin{equation}
    \begin{aligned}
    &\theta^*_f = \arg\max_{\theta} R_f(\theta), \\
    &\theta^*_e = \arg\max_{\theta} R_e(\theta),\\
    &\theta^*_n = \arg\max_{\theta} R_n(\theta).
    \end{aligned}
\end{equation}
Optimizing each reward separately allows each controller to explore the parameter space independently via \( \nabla_\theta R_f(\theta) \), \( \nabla_\theta R_e(\theta) \), and \( \nabla_\theta R_n(\theta) \): the optimizer does not have to compromise between objectives during training, so each controller can learn the structure of its own reward function without interference from the other objectives.

\paragraph{Why not train the controllers jointly?}
Consider instead optimizing a single scalarized (summed) reward:
\begin{equation}
    \begin{aligned}
        \theta^*_{\text{sum}} &= \arg\max_{\theta} R_{\text{sum}}(\theta)\\&=\arg\max_{\theta} (w_n R_n(\theta) + w_f R_f(\theta) + w_e R_e(\theta)).
    \end{aligned}
\end{equation}
The gradient with respect to the parameters \( \theta \) is a weighted sum of the individual gradients:
\begin{equation}
\nabla_\theta R_{\text{sum}}(\theta) = w_f \nabla_\theta R_f(\theta) + w_e \nabla_\theta  R_e(\theta) + w_n \nabla_\theta R_n(\theta).
\end{equation}
The key observation is that this summed gradient may not align with the gradient of any individual reward. If \( \nabla_\theta R_f(\theta) \), \( \nabla_\theta R_e(\theta) \), and \( \nabla_\theta R_n(\theta) \) point in different directions, the optimization step based on the summed gradient does not move in the optimal direction for any individual dimension.

\textit{Example (conflicting gradients).} Suppose at a given point \( \theta \) the gradients are
$\nabla_\theta R_f(\theta) = [1, 0]$, $\nabla_\theta R_e(\theta) = [0, 1]$, and $\nabla_\theta R_n(\theta) = [-1, 0]$. With weights $w_f = w_e = w_n = \frac{1}{3}$, the summed gradient becomes
$\nabla_\theta R_{\text{sum}}(\theta) = [0, \frac{1}{3}]$:
the update moves entirely in the direction of the effectiveness reward, and the feasibility and novelty objectives are not optimized at this step even though all three gradients contributed.

\paragraph{Non-convex Pareto front.}
The set of Pareto-optimal solutions is
\begin{equation}
\begin{aligned}
P =& \{ \theta \in \mathbb{R}^d \mid \nexists \theta' \in \mathbb{R}^d  \\& \text{ such that } R_f(\theta') \geq R_f(\theta), \\& R_e(\theta') \geq R_e(\theta), R_n(\theta') \geq R_n(\theta) \\ &\text{ with at least one strict inequality} \}.
\end{aligned}
\end{equation}
When the objectives are non-linearly dependent, the Pareto front is generally non-convex. Scalarization can only recover Pareto-optimal solutions on the convex hull of the front, and therefore misses solutions in its non-convex regions.
In contrast, training the controllers independently lets each one fully explore the parameter space for its own objective, and the sentence-level decoder can then compose them at inference time to reach trade-off points, including ones that a fixed scalarization cannot express.

\section{Definition of Novelty, Feasibility, and Effectiveness}
\label{appendix:def}

This appendix provides detailed definitions and scoring guidelines for \textbf{Novelty}, \textbf{Feasibility}, and \textbf{Effectiveness}—the three primary dimensions used to evaluate research ideas.

\subsection*{1. Novelty}
Novelty evaluates how different a proposed research idea is compared to existing works. Following previous work \cite{si2024llmsgeneratenovelresearch}, the guidelines for scoring are as follows:

\begin{itemize}
    \item \textbf{1}: \textit{Not novel at all} — The idea is identical to many existing works.
    \item \textbf{3}: \textit{Mostly not novel} — Very similar ideas already exist.
    \item \textbf{5}: \textit{Somewhat novel} — There are differences, but not enough for a standalone paper.
    \item \textbf{6}: \textit{Reasonably novel} — Notable differences, potentially sufficient for a new paper.
    \item \textbf{8}: \textit{Clearly novel} — Major differences from all existing ideas.
    \item \textbf{10}: \textit{Highly novel} — Highly different and creative in a clever, impactful way.
\end{itemize}

\subsection*{2. Feasibility}
Feasibility measures how practical it is to execute the proposed idea within 1–2 months under the following assumptions:
\begin{itemize}
    \item Ample access to OpenAI/Anthropic APIs.
    \item Limited GPU computing resources.
\end{itemize}

Scoring guidelines:
\begin{itemize}
    \item \textbf{1}: \textit{Impossible} — The idea or experiments are fundamentally flawed.
    \item \textbf{3}: \textit{Very challenging} — Major flaws or significant resource limitations.
    \item \textbf{5}: \textit{Moderately feasible} — Possible with careful planning and modifications.
    \item \textbf{6}: \textit{Feasible} — Achievable with reasonable planning.
    \item \textbf{8}: \textit{Highly feasible} — Straightforward to implement and run.
    \item \textbf{10}: \textit{Easy} — Quick to implement without requiring advanced skills.
\end{itemize}

\subsection*{3. Effectiveness}
Effectiveness assesses the likelihood of the research idea achieving meaningful experimental performance improvement. The scoring is defined as:
\begin{itemize}
    \item \textbf{1}: \textit{Extremely unlikely} — Significant flaws, almost certain to fail.
    \item \textbf{3}: \textit{Low effectiveness} — Limited potential, might work in very specific scenarios.
    \item \textbf{5}: \textit{Somewhat ineffective} — A slight chance of marginal or inconsistent improvement.
    \item \textbf{6}: \textit{Somewhat effective} — A decent chance of moderate improvement on certain benchmarks.
    \item \textbf{8}: \textit{Probably effective} — Likely to deliver significant improvement on benchmarks.
    \item \textbf{10}: \textit{Definitely effective} — Highly likely to outperform existing benchmarks by a substantial margin.
\end{itemize}

To ensure reliability, we require the model to provide:
\begin{enumerate}
    \item A brief justification for the score (minimum 2–3 sentences).
    \item References to related works, especially if the score is low.
\end{enumerate}

\section{Comparison with Related Methods}
\label{appendix:compare}

\begin{table}[h!]
\centering
\small
\resizebox{\linewidth}{!}{%

\renewcommand{\arraystretch}{1.2}
\setlength{\tabcolsep}{6pt}
\begin{tabular}{lcccc}
\toprule
\textbf{Method} & \textbf{Task Formulation} & \textbf{Input} & \textbf{Output} & \textbf{Main Focus} \\
\midrule
Ours & Paper + related papers & Main paper + retrieved related & Structured research idea & Open-ended ideation \\
Many Heads Are Better Than One & Task + papers & Research topic + retrieved papers & Structured research idea & Multi-agent ideation \\

AI Scientist & Task + code + Prev Ideas & Task description + experiment.py & Code-constrained idea; experiment plan & End-to-end paper/exp \\
\bottomrule
\end{tabular}}
\caption{Comparison of our method with existing research ideation frameworks.}
\end{table}

AI Scientist focuses on code-based, automated experiment design, which differs from our literature-grounded ideation. 
There are no direct numerical comparisons with AI Scientist.

Many Heads Are Better Than One (MHABTO) is highly relevant as it employs multi-agent LLM discussion for idea generation, evaluated by both LLMs and humans. 
We align our experimental setup with theirs (4 agents, 5 turns). 
As reported in their paper, \textit{Many Heads Are Better Than One} outperforms \textit{AI Scientist}, 
making it a strong reference point for comparison.
% A reference comparison with ChatGPT under the same evaluation protocol is reported in Appendix~\ref{appendix:chatgpt}.

\vspace{5mm}

\section{Comparison with ChatGPT as a Generation Baseline}
\label{appendix:chatgpt}
Our primary objective is to develop a controllable open-source framework to guide smaller models for better research idea generation, which motivates our focus on open models such as T5 and LLaMA2. \\
ChatGPT as a generation baseline is not entirely fair, since it is a significantly larger, proprietary model that is not accessible for training or fine-grained control. Furthermore, in our setup, GPT-4 also serves as the evaluator, which would introduce bias if used as the baseline model. \\
Nevertheless, for reference, we report the result of ChatGPT generations in our evaluation using the same prompts and context inputs as our method with automatic evaluation.
\begin{table}[h]
\small
\centering
\begin{tabular}{lcccc}
\toprule
\textbf{Model} & \textbf{Novelty} & \textbf{Feasibility} & \textbf{Effectiveness} & \textbf{Overall} \\
\midrule
ChatGPT (gpt-4-0314) & 6.2 & 5.3 & 5.4 & 5.6 \\
Ours (LLaMA2-RLHF+Dynamic) & 6.0 & 6.3 & 5.8 & 6.0 \\
\bottomrule
\end{tabular}
\caption{Comparison between ChatGPT and our method with GPT-4 as reviewer.}
\label{tab:chatgpt}
\end{table}\\
As shown in Table \ref{tab:chatgpt}, while ChatGPT achieves a higher novelty score, it tends to over-optimize for novelty at the expense of feasibility and grounding. In contrast, our method produces more balanced and controllable outputs, which we believe are better suited for real-world research ideation workflows.

\newpage
\section{Novelty and Feasibility Control analysis}
\label{appendix:control}
\begin{figure}[h]
    \centering
    \includegraphics[width=0.6\linewidth]{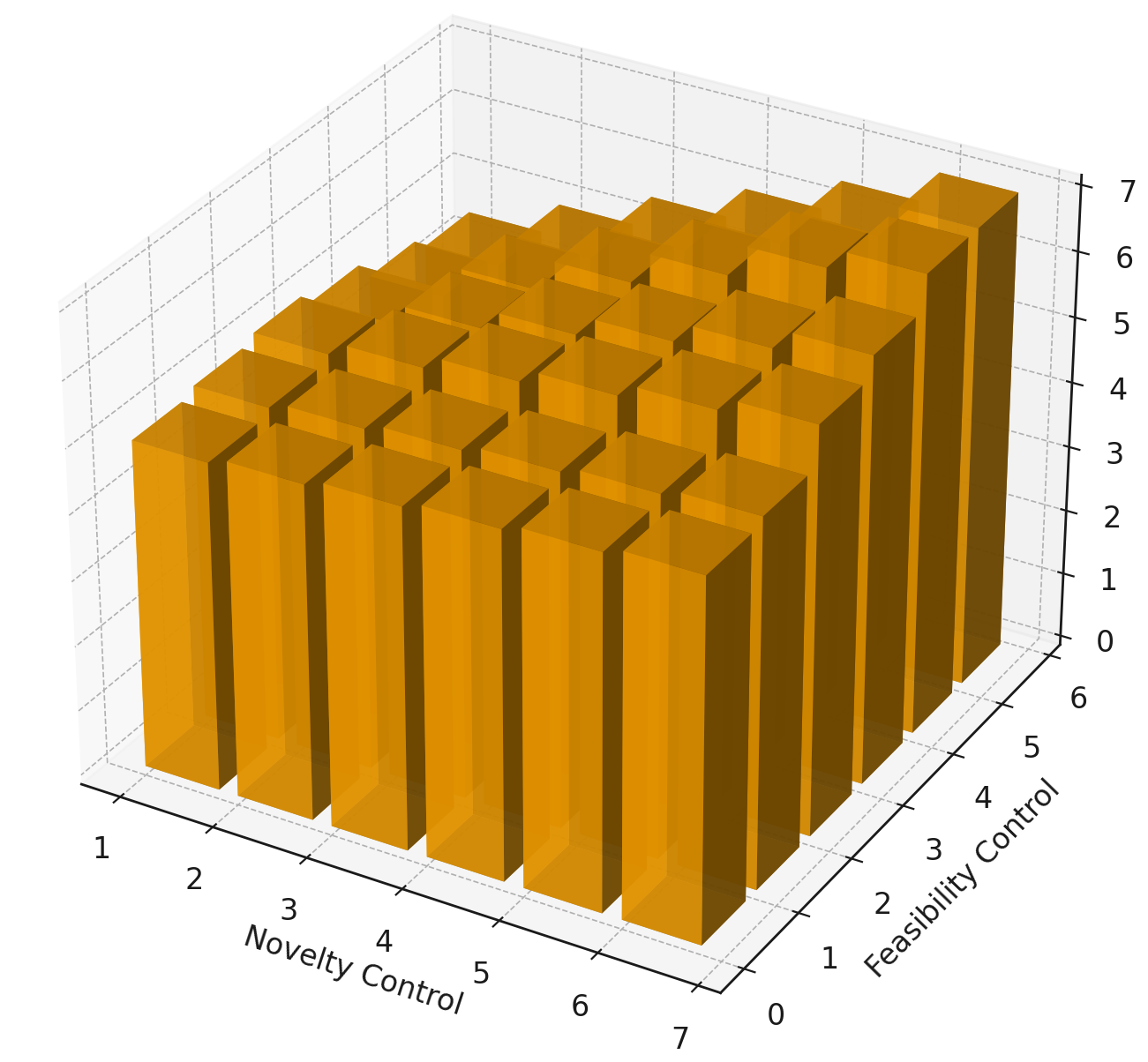}
    \caption{Novelty and Feasibility control analysis}
    \label{fig:control}
\end{figure}
We present the overall score analysis with the control of novelty and feasibility. We can clearly see that with the increase in the control of both dimensions, the overall score increases.
\section{Human Evaluation Barplot}
\label{appendix:humanplot}
\begin{figure}[h]
    \centering
    \includegraphics[width=0.6\linewidth]{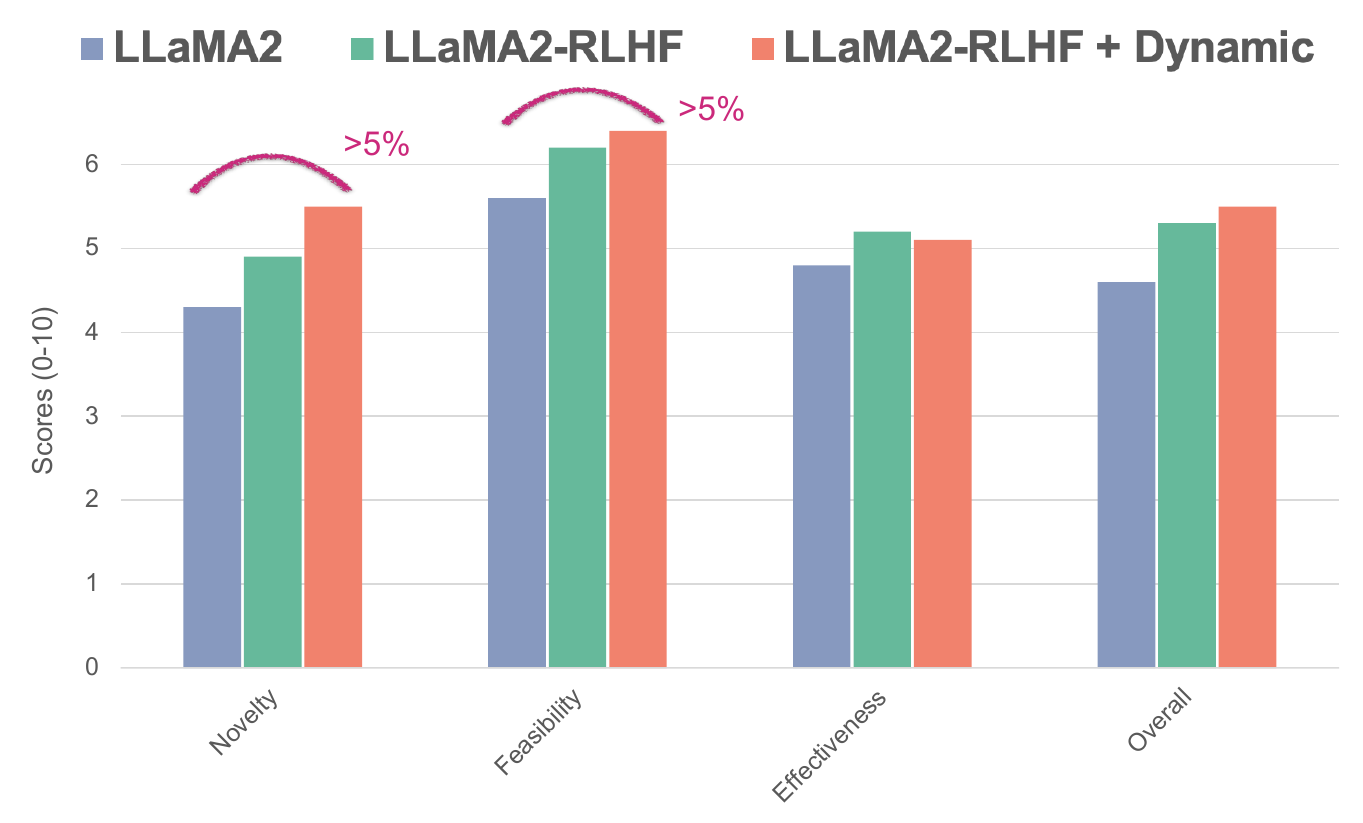}
    \caption{Human Evaluation Results}
    \label{fig:human_eval}
    \vspace{-3mm}
\end{figure}

\section{Scatter of Three Dimension v.s. Overall }
\label{appendix:scatter}
\begin{figure}[h]
    \centering
    \includegraphics[width=0.6\linewidth]{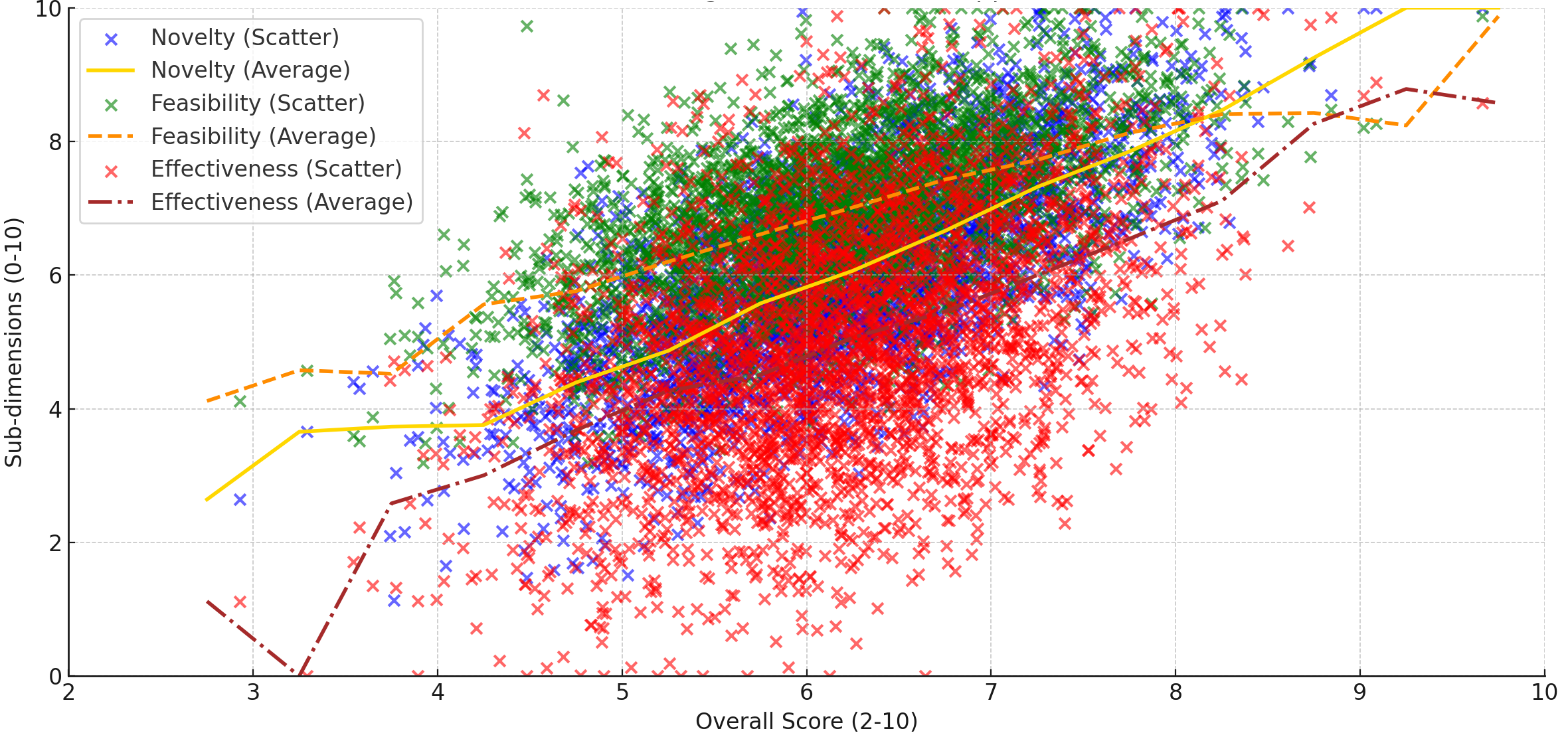}
    \caption{Scatters of different dimensions versus overall scores.}
    \label{fig:scatter}
\end{figure}

\section{Prompt for Research Idea Extraction}
\label{appendix:extract_idea}
\begin{tcolorbox}[appendixbox]
\textbf{System Prompt:} You are an AI assistant whose primary goal is to extract specific details from the scientific literature to aid researchers in understanding and replicating the methodologies and experiment plans of the work.

\subsection*{User Message}
You are tasked with extracting the \textbf{Method} and \textbf{Experiment Plan} from an academic paper. These should include:
\begin{itemize}
    \item \textbf{Method}: A concise summary of the methodological approach employed in the study.
    \item \textbf{Experiment Plan}: Key details of the experiment, including dataset preparation, baseline implementation, and evaluation metrics or procedures.
\end{itemize}
Ensure that the output is clear, focused, and formatted to align with the given structure.

\subsection*{Input Details}
I am going to provide the target paper, related papers, and entities as follows:
\begin{itemize}
    \item \textbf{Target paper title:} \texttt{\{paper['title']\}}
    \item \textbf{Target paper abstract:} \texttt{\{paper['abstract']\}}
    \item \textbf{Entities:} \texttt{\{Entities\}}
\end{itemize}

\subsection*{Objective}
With the provided target paper and entities, extract and summarize the \textbf{Method} and \textbf{Experiment Plan} in the following format:
\begin{itemize}
    \item \textbf{Method:} [Provide a concise description of the methodology used in the study.]
    \item \textbf{Experiment Plan:} [Summarize the dataset preparation, baseline implementation, and evaluation procedures.]
\end{itemize}

\subsection*{Example Input}
\begin{itemize}
    \item \textbf{Target paper title:} "Transformer Models for Legal Text Analysis"
    \item \textbf{Target paper abstract:}
    \begin{quote}
        "Deep learning has transformed the field of natural language processing, yet challenges remain in domain-specific applications. This paper explores the use of transformer models for legal text analysis, addressing the question: 'Can pre-trained language models be adapted effectively for legal case prediction?' The study employs fine-tuning techniques and evaluates performance on a benchmark dataset of legal cases. Results show a significant improvement in prediction accuracy compared to traditional methods."
    \end{quote}
\end{itemize}

\subsection*{Expected Output}
\begin{itemize}
    \item \textbf{Method:} We introduce fine-tuning techniques to adapt pre-trained transformer models for legal text analysis.
    \item \textbf{Experiment Plan:} 
    \begin{itemize}
        \item \textbf{Dataset Preparation:} A legal benchmark dataset of case documents is used.
        \item \textbf{Baseline Implementation:} Models are compared against traditional NLP methods.
        \item \textbf{Evaluation Procedure:} Performance is measured in terms of prediction accuracy on unseen legal cases.
    \end{itemize}
\end{itemize}
\end{tcolorbox}

\section{Prompt for Novelty Score Extraction}
\label{appendix:prompt-nov}
\begin{tcolorbox}[appendixbox]
\noindent
\textbf{System Prompt:}
You are a specialized assistant for scientific text evaluation. Your task is to evaluate the novelty of scientific papers.

\subsection*{User Prompt}
Based on the following information about a scientific paper, please evaluate its novelty:

\begin{itemize}
    \item \textbf{Title:} \texttt{\{title\}}
    \item \textbf{Abstract:} \texttt{\{abstract\}}
    \item \textbf{Related Works (top 3 from citations since 2023):} \texttt{\{recent\_works\}}
    \item \textbf{Review Comments:} \texttt{\{reviews\}}
\end{itemize}

\subsection*{Novelty Evaluation Prompt}
\label{appendix:extract_nov}
Evaluate how creative and different the idea is compared to existing works on the topic. Consider all papers that appeared online prior to July 2024 as existing work. Your evaluation should consider the degree to which the paper brings new insights and differentiates itself from prior research.

\textbf{Scoring Criteria:}\\
Please assign a novelty score on a scale from 1 to 10 based on the following criteria:\\
\textbf{Novelty Definition:}\\
We score the novelty of papers based on how different they are from existing works. The guidelines for scoring novelty are:
\begin{itemize}
    \item \textbf{1:} Not novel at all — many existing ideas are the same.
    \item \textbf{3:} Mostly not novel — very similar ideas exist.
    \item \textbf{5:} Somewhat novel — differences exist but not enough for a new paper.
    \item \textbf{6:} Reasonably novel — notable differences, could lead to a new paper.
    \item \textbf{8:} Clearly novel — major differences from all existing ideas.
    \item \textbf{10:} Very novel — highly different and creative in a clever way.
\end{itemize}

\textbf{Novelty Rationale:}\\
After assigning a score, provide a short justification for your rating. If the score is below 6, specify similar works that closely resemble this paper. The rationale should be at least 2-3 sentences.

\textbf{Output Format:}\\
The result must be output in JSON format, as shown in the example below:
\begin{quote}
\texttt{\{"score": 8, "reason": "This paper introduces a novel machine learning approach for earthquake prediction using real-time seismic data, which represents a significant improvement over traditional statistical models. By incorporating both real-time data and deep learning techniques, this approach enables more accurate and timely earthquake forecasts. Although there are existing works using machine learning for seismic analysis, the integration of real-time data and advanced neural networks distinguishes this paper. The comprehensive validation of the method, including comparisons with conventional models, highlights its contribution to the field."\}}
\end{quote}

\noindent
The response should \textbf{only contain JSON content}.
\end{tcolorbox}

\section{Prompt for Research Idea Generation}
\label{appendix:gen_idea}
\begin{tcolorbox}[appendixbox]

\textbf{System Prompt:}  
You are an AI assistant specializing in extracting and generating structured research ideas from scientific papers. Your task is to assist researchers in developing concise, clear, and innovative research ideas based on the provided input.  

\vspace{0.3em}  
\textbf{User Instructions:}  
You are tasked with generating a structured research idea that includes:  
\begin{itemize}
    \item \textbf{Method:} A concise summary of the methodological approach employed in the study.  
    \item \textbf{Experiment Plan:} Key details of the experiment, including dataset preparation, baseline implementation, and evaluation procedures.
    \item \textbf{Problem:} A clear statement of the research problem or gap the study aims to address.
    \item \textbf{Related Works:} Identify and summarize the top 3 most relevant related works, emphasizing how the target paper builds upon or differs from them.
\end{itemize}  

Ensure that the output adheres to the following requirements:  
\begin{enumerate}
    \item \textbf{Contextual Relevance:} The generated idea must align with the main theme of the provided paper and incorporate any specified entities or constraints.  
    \item \textbf{Clarity and Structure:} The output must be structured, clear, and concise, formatted as follows:  
    \begin{quote}
        \textbf{Problem:} [Description of the research problem or gap being addressed.]  
        
        \textbf{Method:} [Concise description of the methodology used in the study.]  
        
        \textbf{Experiment Plan:}  
        \begin{itemize}
            \item Dataset Preparation: [Details of the dataset used.]  
            \item Baseline Implementation: [Details of the baseline setup.]  
            \item Evaluation Procedure: [Evaluation metrics and procedures used.]  
        \end{itemize}
        
        \textbf{Related Works:}  
        \begin{itemize}
            \item \textbf{Work 1:} [Summary of the first related work.]  
            \item \textbf{Work 2:} [Summary of the second related work.]  
            \item \textbf{Work 3:} [Summary of the third related work.]  
        \end{itemize}
    \end{quote}  
\end{enumerate}  

\textbf{Example Input:}  
\begin{itemize}
    \item \textbf{Target Paper Title:} "Transformer Models for Legal Text Analysis"  
    \item \textbf{Abstract:} "This study explores fine-tuning transformer models for legal text analysis and evaluates their performance on a benchmark dataset, achieving significant accuracy improvements over traditional methods."  
    \item \textbf{Problem:} Traditional NLP methods often fail to capture the complex linguistic structure and contextual dependencies in legal text, leading to suboptimal accuracy in legal text analysis tasks.
    \item \textbf{Entities:} Legal datasets, transformer models, benchmark evaluation.  
    \item \textbf{Related Works:}  
    \begin{itemize}
        \item Work 1: "BERT for Legal Case Prediction" focuses on fine-tuning BERT models for legal document classification.  
        \item Work 2: "Legal NLP with Statistical Models" applies traditional NLP techniques for legal text analysis.  
        \item Work 3: "Adapting Transformers for Domain-Specific Tasks" investigates transformer models in specialized fields like healthcare and law.  
    \end{itemize}
\end{itemize}  

\textbf{Example Output:}  
\begin{quote}
    \textbf{Problem:} Traditional NLP methods often fail to capture the complex linguistic structure and contextual dependencies in legal text, leading to suboptimal accuracy in legal text analysis tasks.  
    
    \textbf{Method:} We introduce fine-tuning techniques to adapt pre-trained transformer models for legal text analysis, focusing on improved generalization.  
    
    \textbf{Experiment Plan:}  
    \begin{itemize}
        \item \textbf{Dataset Preparation:} A benchmark dataset of legal case documents is pre-processed and tokenized.  
        \item \textbf{Baseline Implementation:} Traditional NLP methods are used as the baseline for comparison.  
        \item \textbf{Evaluation Procedure:} Prediction accuracy is measured on unseen legal cases using cross-validation techniques.  
    \end{itemize}  
    
    \textbf{Related Works:}  
    \begin{itemize}
        \item \textbf{Work 1:} "BERT for Legal Case Prediction" explores fine-tuning BERT for classification, but lacks transformer-level insights specific to domain challenges.  
        \item \textbf{Work 2:} "Legal NLP with Statistical Models" applies rule-based methods but achieves lower accuracy and generalizability compared to transformer models.  
        \item \textbf{Work 3:} "Adapting Transformers for Domain-Specific Tasks" provides foundational techniques but does not address challenges in legal text structure.  
    \end{itemize}
\end{quote}
\end{tcolorbox}

\section{Prompt for Automatic Evaluation}
\label{appendix:auto_eval}
\begin{tcolorbox}[appendixbox]

\textbf{System Prompt:}  
You are an AI reviewer specializing in evaluating the quality of research ideas based on specific criteria: \textbf{Novelty}, \textbf{Feasibility}, and \textbf{Effectiveness}. Your task is to assess each criterion and provide structured feedback for automatic evaluation.

\vspace{0.3em}
\textbf{User Instructions:}  
For a given research idea, evaluate the following dimensions:  

\begin{enumerate}
    \item \textbf{Novelty:} Assess how creative and unique the idea is compared to existing works.  
    \item \textbf{Feasibility:} Evaluate the practicality of executing the idea within typical resource constraints.  
    \item \textbf{Effectiveness:} Judge the potential of the idea to achieve its intended objectives or performance improvements.  
\end{enumerate}

\textbf{Scoring Criteria:}  
Provide a score between 1 and 10 for each dimension, adhering to these guidelines:  
{\{Add detailed definition of 3 Metrics HERE\}}

\textbf{Evaluation Output Requirements:}  
Provide a structured evaluation as follows:  

\begin{itemize}
    \item Score for each dimension (\textbf{Novelty}, \textbf{Feasibility}, \textbf{Effectiveness}).  
    \item Brief justification (minimum 2–3 sentences) for each score.  
    \item If the score is below 6, include references to related works or specific reasons for the low rating.  
\end{itemize}

\textbf{Example Input:}  
\begin{itemize}
    \item \textbf{Title:} "Transformer Models for Legal Text Analysis"  
    \item \textbf{Abstract:} "This paper explores fine-tuning transformer models for legal text analysis, demonstrating significant accuracy improvements over traditional methods."  
    \item \textbf{Generated Idea:}  
    \begin{quote}
        Method: Fine-tune pre-trained transformer models for legal case prediction.  
        Experiment Plan: Use a benchmark legal dataset, traditional NLP methods as baselines, and evaluate using prediction accuracy.  
    \end{quote}
\end{itemize}

\textbf{Example Output:}  
\begin{quote}
\{
    \texttt{"novelty": 8,} \\
    \texttt{"novelty\_justification": "The idea introduces transformer-based approaches to legal text analysis, offering a clear improvement over rule-based and statistical methods.",} \\
    \texttt{"feasibility": 6,} \\
    \texttt{"feasibility\_justification": "Implementation is feasible with access to pre-trained models and benchmark datasets, though computational cost may be a concern.",} \\
    \texttt{"effectiveness": 7,} \\
    \texttt{"effectiveness\_justification": "The method has a high likelihood of outperforming traditional baselines based on prior research in similar domains."} \\
\}
\end{quote}
\end{tcolorbox}

\end{document}